\documentclass[10pt,twocolumn,letterpaper]{article}

\usepackage{iccv}
\usepackage{times}
\usepackage{epsfig}
\usepackage{graphicx}
\usepackage{amsmath}
\usepackage{amssymb}
\usepackage{multicol}
\usepackage{array}
\usepackage{float}

\usepackage{booktabs}
\usepackage{multirow}
\usepackage{tabu}
\usepackage{makecell}
\usepackage[titletoc,toc,title]{appendix}

\DeclareMathOperator*{\argmaxC}{\arg\max}   

\usepackage[pagebackref=true,breaklinks=true,letterpaper=true,colorlinks,bookmarks=false]{hyperref}

\iccvfinalcopy 

\def\iccvPaperID{3410} 

\ificcvfinal\fi
\begin{document}

\title{AdaptIS: Adaptive Instance Selection Network}

\author{Konstantin Sofiiuk\\
Samsung AI Center\\
{\tt\small k.sofiiuk@samsung.com}
\and
Olga Barinova\\
Samsung AI Center\\
{\tt\small o.barinova@samsung.com}
\and
Anton Konushin\\
Samsung AI Center\\
{\tt\small a.konushin@samsung.com}
}

\maketitle

\begin{abstract}
We present \emph{Adaptive Instance Selection} network architecture for class-agnostic instance segmentation. Given an input image and a point $(x, y)$, it generates a mask for the object located at $(x, y)$. The network adapts to the input point with a help of AdaIN layers \cite{karras2018style}, thus producing different masks for different objects on the same image. AdaptIS generates pixel-accurate object masks, therefore it accurately segments objects of complex shape or severely occluded ones. AdaptIS can be easily combined with standard semantic segmentation pipeline to perform panoptic segmentation. To illustrate the idea, we perform experiments on a challenging toy problem with difficult occlusions. Then we extensively evaluate the method on panoptic segmentation benchmarks. We obtain state-of-the-art results on Cityscapes and Mapillary even without pretraining on COCO, and show competitive results on a challenging COCO dataset. The source code of the method and the trained models are available at  \href{https://github.com/saic-vul/adaptis}{https://github.com/saic-vul/adaptis}.
\end{abstract}

\section{Introduction}
While humans can easily segment objects in images, this task is quite difficult to formalize. One of the formulations is \emph{semantic segmentation} which aims at assigning a class label to each image pixel. Another formulation is \emph{instance segmentation}, which aims to detect and segment each object instance in the image. While semantic segmentation does not imply separating objects of the same class, instance segmentation focuses only on segmenting things (\ie countable objects such as people, animals, tools) and does not account for so-called stuff (\ie amorphous  regions  of  similar texture or material such as grass, sky, road). Recently, \emph{panoptic segmentation} task that combines semantic and instance segmentation together, has been introduced \cite{kirillov2018panoptic}. The goal of panoptic segmentation is to map each pixel of an image to a pair $(l_i, z_i) \in L \times N$,  where $l_i$ represents the semantic class of pixel $i$ and $z_i$ represents its instance id.

\begin{figure}[t]
    \includegraphics[width=\linewidth]{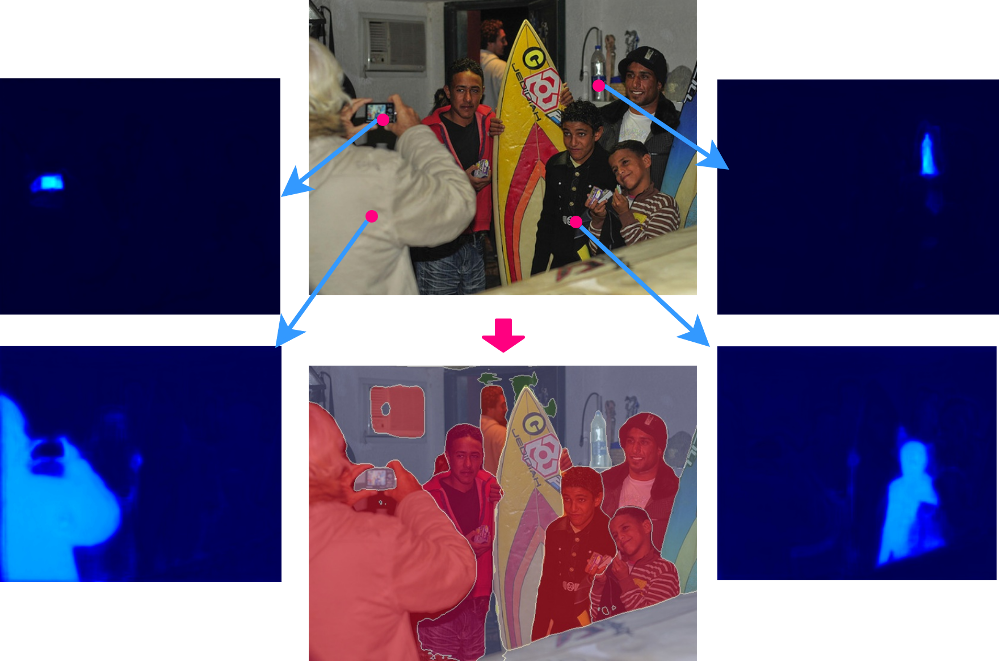}
    \caption{AdaptIS takes an image and a point proposal $(x, y)$ as input and generates a mask for an object located at position $(x, y)$. AdaptIS can be easily combined with standard semantic segmentation pipeline to perform panoptic segmentation. The picture shows sampled point proposals, corresponding generated masks, and the final result of panoptic segmentation.}
    \label{fig:motivation}
\end{figure}

In this work we focus on instance and panoptic segmentation. We introduce AdaptIS, a fully differentiable end-to-end network for class-agnostic instance segmentation. The network takes an image and a point proposal $(x, y)$ as input and generates a mask of an object located at position $(x, y)$. Figure \ref{fig:motivation} shows examples of the masks produced by AdaptIS for different input point proposals. 

\begin{figure}
    \includegraphics[width=\linewidth]{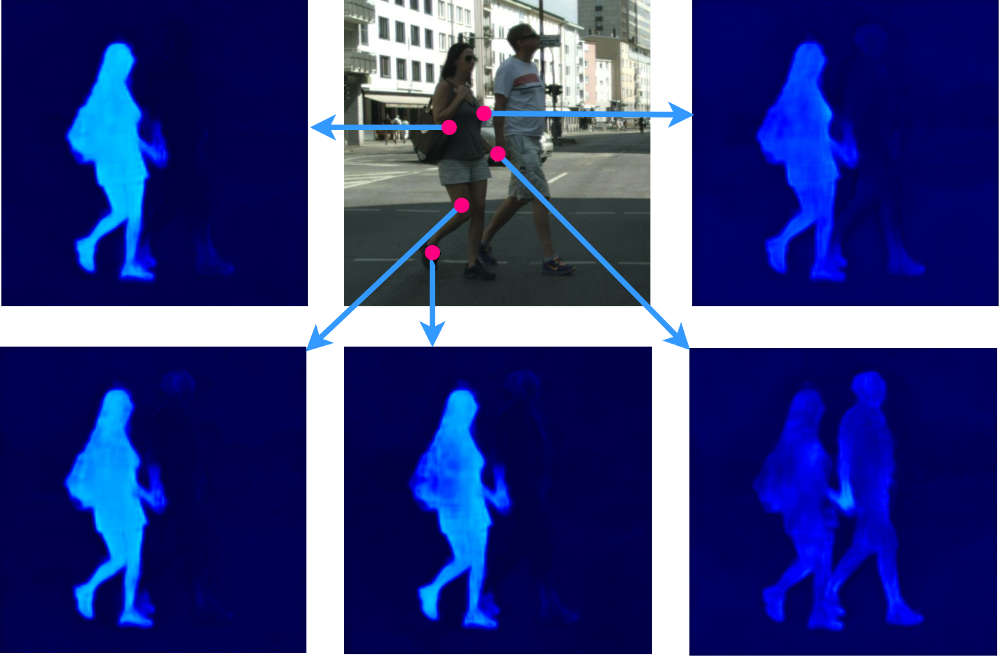}
    \caption{Outputs of AdaptIS for different point proposals. For most points that belong to the same object AdaptIS by design produces very similar masks. Note that the point on the border of two objects (two persons walking hand in hand) results in a merged mask of the two objects.}
    \label{fig:point_proposals}
\end{figure}

We cast the problem of generating an object mask as binary segmentatiIn contrast to the mainstream detection-first methods for instance segmentation \cite{he2017mask, liu2018path} AdaptIS does not rely on bounding box proposals. Instead, it directly optimizes target segmentation accuracy. In contrast to instance embedding-based methods \cite{fathi2017semantic, de2017semantic, novotny2018semi} which map the pixels of an image into embedding space and then perform clustering, AdaptIS does not require heuristic post-processing steps. Instead, it directly generates an object mask for a given point proposal.
on. For a given image $I$ and a fixed point proposal $(x, y)$ we simply optimize just one target loss function. We use a pixel-wise loss comparing AdaptIS prediction to the mask of the target object located at position $(x, y)$ on the image $I$. At train time for each object we sample different point proposals. Thus, the network learns to generate the same object mask given different points on the same object (see Figure \ref{fig:point_proposals}). Interestingly, AdaptIS does not use semantic labels during training, which allows it to segment objects of any class. 

Proposed approach has many practical advantages over existing methods. Compared to detection-first methods, AdaptIS is much better suited for the cases of severe occlusions. Since it provides pixel-accurate segmentation and can better handle objects of complex shape. Besides, AdaptIS can segment objects of different size using only single-scale features. Due to its simplicity AdaptIS has very few train-time hyperparameters and therefore can be trained on a new dataset almost without fine-tuning. AdaptIS can be easily paired with standard semantic segmentation pipeline to provide panoptic segmentation. Examples of panoptic segmentation results for Cityscapes are shown in Figure \ref{fig:AdaptIS_properties}.

In the next sections we discuss related work and explain proposed architecture for instance segmentation. Then we describe an extension of the model to provides semantic labels and propose a greedy algorithm for panoptic segmentation using AdaptIS. We perform experiments on a toy problem with difficult occlusions which is extremely challenging for the detection-first segmentation methods like Mask R-CNN \cite{he2017mask}. In the end, we evaluate the method on Cityscapes, Mapillary and COCO benchmarks. We obtain state-of-the-art accuracy on Cityscapes and Mapillary and demonstrate competitive performance on COCO.

\begingroup
\setlength{\tabcolsep}{0.8pt} 
\renewcommand{\arraystretch}{0.5} 
\begin{figure}[t]
 \begin{tabular}{cccccccc}
    \includegraphics[width=0.19\linewidth]{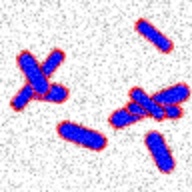} &
    \includegraphics[width=0.19\linewidth]{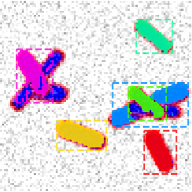} &
    \includegraphics[width=0.19\linewidth]{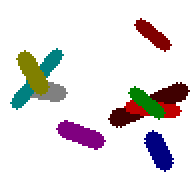} &
    \multirow{2}{*}[1.4cm]{\includegraphics[height=0.41\linewidth]{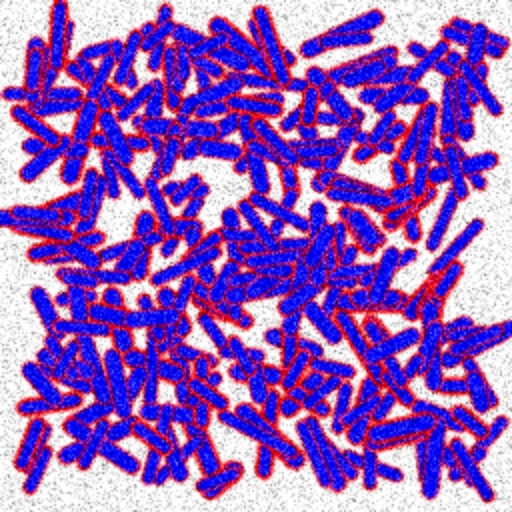}} \\

    \includegraphics[width=0.19\linewidth]{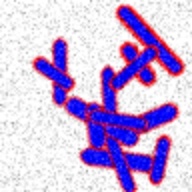} &
    \includegraphics[width=0.19\linewidth]{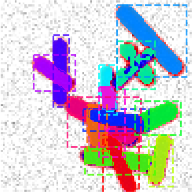} &
    \includegraphics[width=0.19\linewidth]{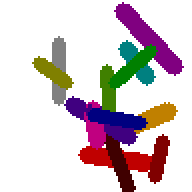} &
    \\
    \includegraphics[width=0.19\linewidth]{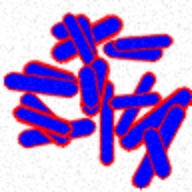} &
    \includegraphics[width=0.19\linewidth]{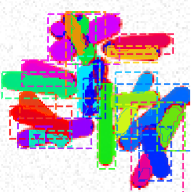} &
    \includegraphics[width=0.19\linewidth]{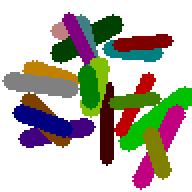} &
    \multirow{2}{*}[1.4cm]{\includegraphics[height=0.41\linewidth]{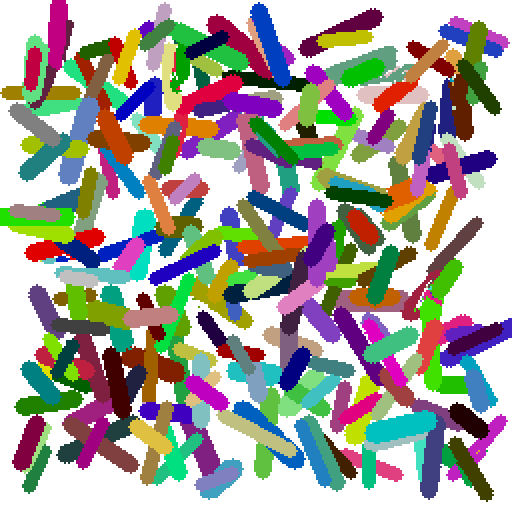}} 
    \\
    \includegraphics[width=0.19\linewidth]{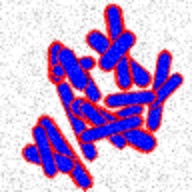} &
    \includegraphics[width=0.19\linewidth]{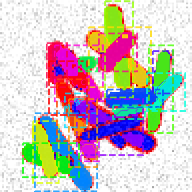} & 
    \includegraphics[width=0.19\linewidth]{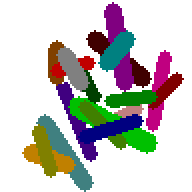} &
    \\
    (a) & (b) & (c) & (d)
  \end{tabular}
  \caption{Results of class-agnostic instance segmentation for the toy data: (a) --- validation images, (b) --- results of Mask R-CNN, (c) --- results of AdaptIS. It should be noted that Mask R-CNN often fails in cases when objects significantly overlap. (d) --- larger image (top) and result of instance segmentation with AdaptIS (bottom). Though AdaptIS was trained on smaller images like shown in column (a), it works quite well on larger images like the one shown in (d). In this example it has correctly segmented 234 of 250 objects, missed 16 objects and made just 3 false positives.}
  \label{fig:toy_examples}
\end{figure}
\endgroup

\section{Related work}

\textbf{Detection-first instance segmentation methods.} 
In recent years the approach to instance segmentation based on standard object detection pipeline \cite{he2017mask, chen2018masklab, liu2018path} became a mainstream, achieving state-of-the-art results on various benchmarks. However this detection-first approach has a few severe limitations. First, strongly overlapping objects may have very similar bounding boxes or even share the same bounding box. In this case, the mask branch of the network has no information about which object to segment inside a bounding box. Second, the process of mask inference is based on ROI pooling operation. ROI pooling reduces dimensionality of object features, which leads to loss of information and causes inaccurate masks for objects of complex shape.

\textbf{Instance embedding methods.} 
Another interesting approach to instance segmentation is based on instance embedding \cite{de2017semantic, fathi2017semantic, kong2018recurrent}. In this approach a network maps each pixel of an input image into an embedding space. The network learns to map the pixels of the same object into close points of the embedding space, while pixels that belong to different objects are mapped into distant points of the embedding space. This property of the network is enforced by introducing several losses that directly control proximity of the embeddings. After computing the embeddings one needs to obtain instance segmentation map using some clustering method. In contrast to detection-first methods this approach allows for producing pixel-accurate object masks. However, it has it's own limitations. Usually, some heuristic clustering procedures are applied to the resulting embeddings, \eg mean shift \cite{de2017semantic} or region growing \cite{fathi2017semantic}. In many cases this leads to sub-optimal results, as the network is not optimizing segmentation accuracy. Currently, the methods of this group demonstrate inferior performance on standard benchmarks compared to detection-first methods.

\textbf{Panoptic segmentation methods.}
The formulation of panoptic segmentation problem was first introduced in \cite{kirillov2018panoptic}. In this work the authors presented a strong baseline that combines two independent networks for semantic \cite{zhao2017pyramid} and instance segmentation \cite{he2017mask} followed by heuristic post-processing and late fusion of the two outputs. Later the same authors proposed a single model based on feature pyramid network that showed higher accuracy than the combination of two networks \cite{kirillov2019panoptic}. The other methods for panoptic segmentation base on standard instance or semantic segmentation pipelines and usually introduce additional losses that enforce some heuristic constraints on segmentation consistency \cite{li2018learning, yang2019deeperlab}.

\section{Architecture overview}

\begin{figure}[t]
    \includegraphics[width=\linewidth]{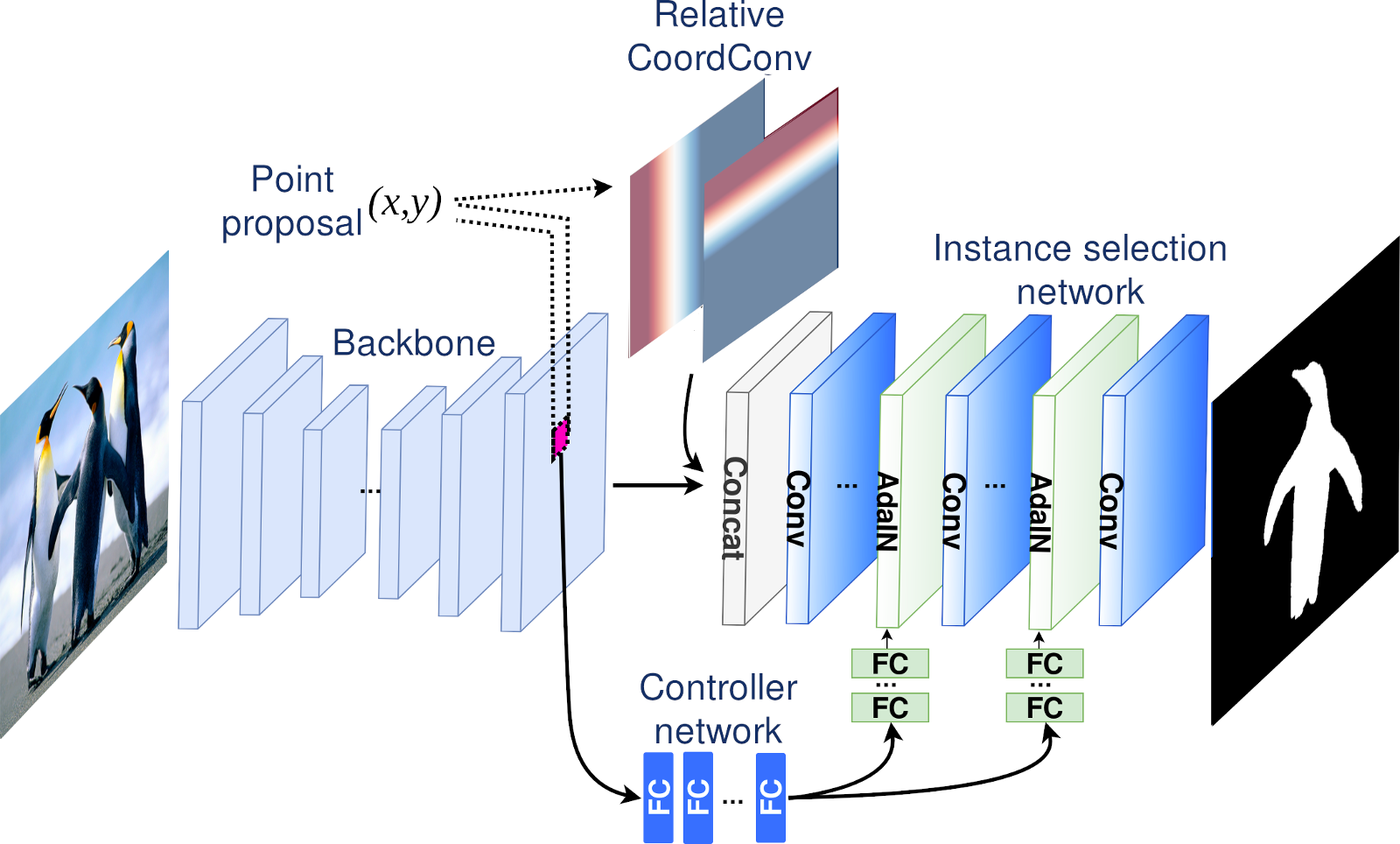}
    \caption{Architecture of AdaptIS for class-agnostic instance segmentation. The network is built on top of a (pretrained) \emph{backbone} for feature extraction. It introduces 1) a lightweight \emph{instance selection network} for inferring object masks that adapts to a \emph{point proposal} using AdaIN mechanism \cite{huang2017arbitrary, dumoulin2017learned, ghiasi2017exploring, karras2018style}; 2) a \emph{controller network} that fetches the feature at position $(x, y)$, processes it in a series of fully connected layers, and provides an input for AdaIN layers in instance selection network; 3) \emph{Relative CoordConv} block that helps to disambiguate similar objects located at different positions in the image.}
    \label{fig:explain}
\end{figure}

Schematic structure of the AdaptIS is shown in Figure \ref{fig:explain}. The network takes an image and a point proposal $(x, y)$ as input and outputs a mask of an object located at position $(x, y)$. Below we explain the main components of the architecture.

\textbf{Instance selection network.}
Let us have a closer look at the generation of an object mask. Suppose that we are given a vector that characterizes some object allowing us to distinguish it from other objects in the image.

But how can we generate an object mask based on this ''characteristic'' vector? Recently an elegant solution for similar problems was proposed in the literature on style transfer \cite{huang2017arbitrary, dumoulin2017learned, ghiasi2017exploring} and image generation \cite{karras2018style}. Using Adaptive Instance Normalization (AdaIN) layers \cite{huang2017arbitrary}, one can parameterize a network, \ie vary the network output for the same input by providing different parameters to AdaIN.

In this work we propose a lightweight \emph{Instance Selection Network}, which is parameterized using AdaIN. To the best of our knowledge, adaptive network architectures have not been used for object segmentation previously. ''Characteristic'' vector thus contains parameters for AdaIN layers of the instance selection network. Instance selection network generates an object mask using features extracted by a backbone and a ''characteristic'' vector.

\textbf{Point proposals and controller network.} 
The next question is: where can we get a good ''characteristic'' vector for an object? 

Let us recall the idea behind the instance embedding methods \cite{de2017semantic, fathi2017semantic, kong2018recurrent, novotny2018semi}. In these methods, a network maps image pixels into embedding space, pushing embeddings closer for the pixels of the same object and pulling them apart for the pixels of different objects. Thus, an embedding $Q(x, y)$ of a pixel at position $(x, y)$ can be viewed as unique features of an object located at $(x, y)$. We exploit this idea in AdaptIS architecture and generate a ''characteristic'' vector from the embedding $Q(x, y)$. 

The output of a backbone $Q$ may have lower spatial resolution compared to the input image. Thus, for each point $(x, y)$ we obtain corresponding embedding $Q(x, y)$ with the use of bilinear interpolation. The resulting embedding $Q(x, y)$ is then processed in a \emph{controller network}, that consists of a few fully connected layers. Controller network outputs a ''characteristic'' vector, which is then passed to AdaIN layers in instance selection network. Using this mechanism the instance selection network adapts to the selected object.

Note, that in contrast to instance embedding methods we do not optimize the distances between embeddings. Instead, the controller network connects backbone and instance selection network in a feedback loop, thus enforcing backbone to produce rich features characterizing each particular object. As a result, AdaptIS directly optimizes accuracy of an object mask without a need for auxiliary losses.

\textbf{Relative CoordConv.}
\cite{novotny2018semi} showed that one necessarily needs to provide pixel coordinates to the network in order to disambiguate different object instances. \cite{liu2018intriguing} proposed a simple CoordConv layer that feeds pixel coordinates to the network. It creates a tensor of same size as input that contains pixel coordinates normalized to $[-1, 1]$. This tensor is then concatenated to the input features and passed to the next layers of the network. 

However, in AdaptIS we also need to provide the coordinates of an object to the network. Moreover, we would prefer not to use CoordConv in backbone, as standard backbones are pre-trained without it. Instead, we propose to use a \emph{Relative CoordConv} block that depends on the point proposal and is used only in the later layers of AdaptIS. Similarly to original CoordConv, Relative CoordConv produces two maps, one for $x$-coordinates and one for $y$-coordinates. The values of each map vary from $-1$ to $+1$, where $-1$ corresponds to $x-R$ (or $y-R$) and $+1$ corresponds to $x+R$ (or $y+R$). $R$ is a radius of Relative CoordConv that is a hyperparameter of the algorithm (it roughly sets the maximum size of an object). One may view the Relative CoordConv block as a prior on the location of an object. The output of Relative CoordConv is concatenated with the features produced by a backbone, and together they are passed to an instance selection network. Similar blocks have been used in the literature, \eg in \cite{ganin2016deepwarp}.

\begingroup
\begin{centering}
\setlength{\tabcolsep}{0.8pt} 
\renewcommand{\arraystretch}{0.8} 
\begin{figure}[t]
 \begin{tabular}{ccccc}
    \includegraphics[width=0.19\linewidth]{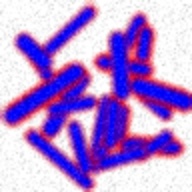} &
    \includegraphics[width=0.19\linewidth]{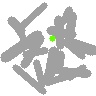} &
    \includegraphics[width=0.19\linewidth]{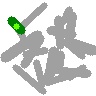} &
    \includegraphics[width=0.19\linewidth]{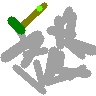} &
    \includegraphics[width=0.19\linewidth]{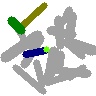} \\
    \includegraphics[width=0.19\linewidth]{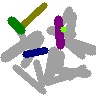} &
    \includegraphics[width=0.19\linewidth]{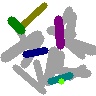} &
    \includegraphics[width=0.19\linewidth]{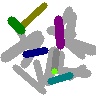} &
    \includegraphics[width=0.19\linewidth]{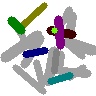} &
    \includegraphics[width=0.19\linewidth]{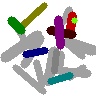} \\
    \multicolumn{1}{c}{...} &
    \includegraphics[width=0.19\linewidth]{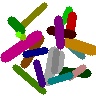} &
    \includegraphics[width=0.19\linewidth]{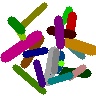} &
    \includegraphics[width=0.19\linewidth]{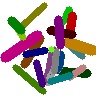} &
    \includegraphics[width=0.19\linewidth]{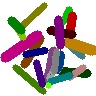} \\
  \end{tabular}
  \caption{Instance segmentation with AdaptIS. Point proposals at each iteration of the method are shown in light-green. Different objects are marked with different colors. ''Unknown'' areas of the images are shown in grey. At each iteration a new point proposal is sampled and a corresponding object is segmented. See text for more details.}
  \label{fig:toy_instance_segmentation}
\end{figure}
\end{centering}
\endgroup

\textbf{Training.}
At train time for each image in a batch we sample $K$ points $(x_i, y_i), i=1 \dots K$ using object-level stratification. We pick a random object and then sample a random point at that object.

Having sampled the features $Q(x_i, y_i), i=1 \dots K$, we train a network to predict corresponding object masks $M_i, i=1 \dots K$ from an input image $I$. In order to do that that we optimize a pixel-wise loss function that compares network output for a pair $\big(I, Q(x_i, y_i)\big)$ to the ground truth object mask $M_i$. Generally, binary cross entropy is used for this type of problems. However, in our experiments we used a modification of Focal Loss \cite{lin2017focal} that demonstrated slightly better results (see appendix for more details). The gradient from the loss propagates to all components of AdaptIS and to the backbone.

Random sampling of the point proposals during training automatically enforces AdaptIS to produce similar output for point proposals that belong to the same object at test time. In contrast to instance embedding methods that directly optimize the distances between the embeddings we directly optimize the resulting object masks and provide only soft constraints on the embeddings by the similarity of the masks. Examples of the learned masks for different point proposals are shown in Figure \ref{fig:point_proposals}.

\section{Class-agnostic instance segmentation}
\label{sec:inference}

\begingroup
\setlength{\tabcolsep}{1pt} 
\renewcommand{\arraystretch}{0.5} 
\begin{figure*}[t]
 \begin{tabular}{ccc}
    \includegraphics[width=0.33\linewidth]{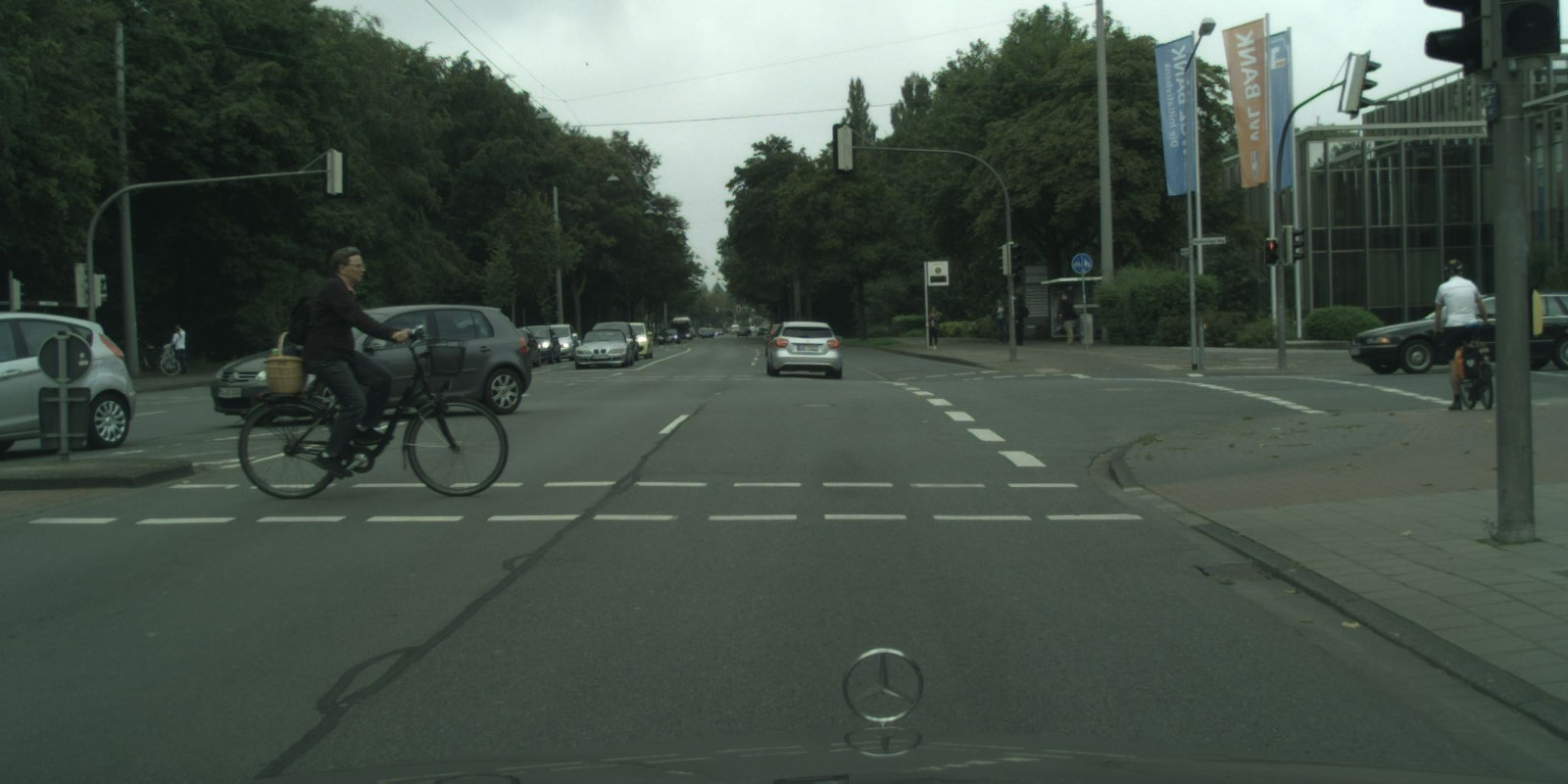} &
    \includegraphics[width=0.33\linewidth]{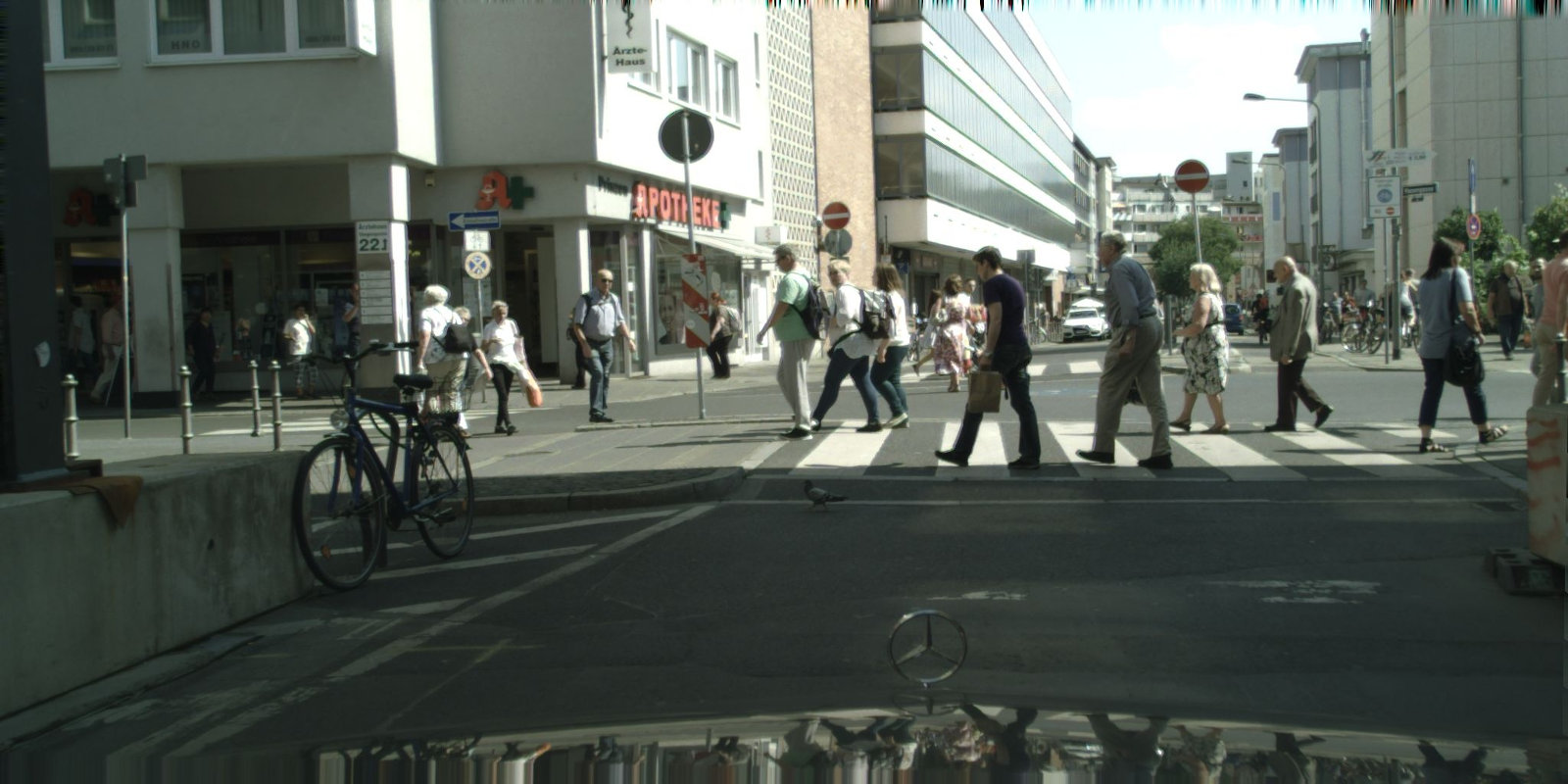} &
    \includegraphics[width=0.33\linewidth]{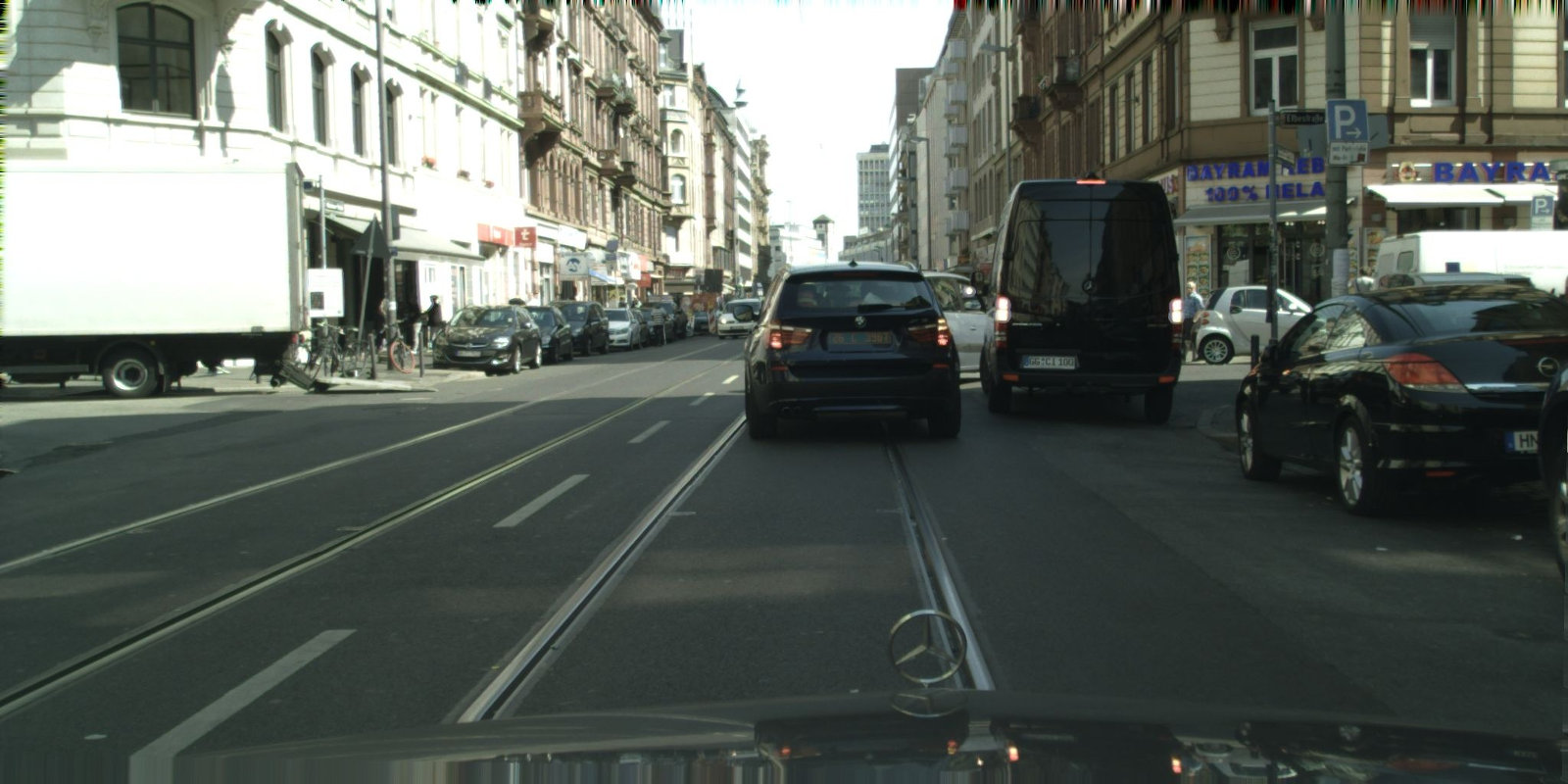} \\
    \includegraphics[width=0.33\linewidth]{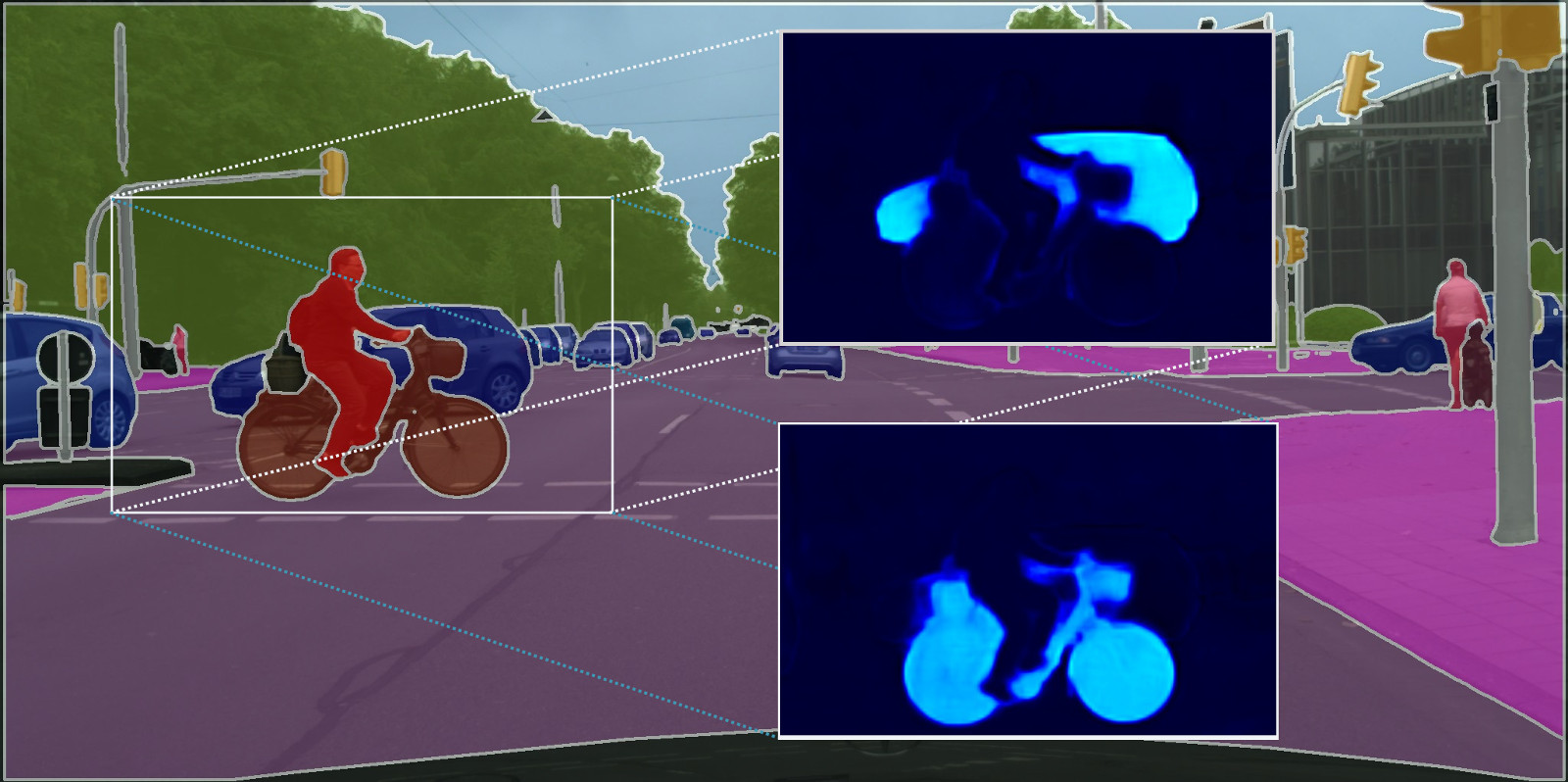} &
    \includegraphics[width=0.33\linewidth]{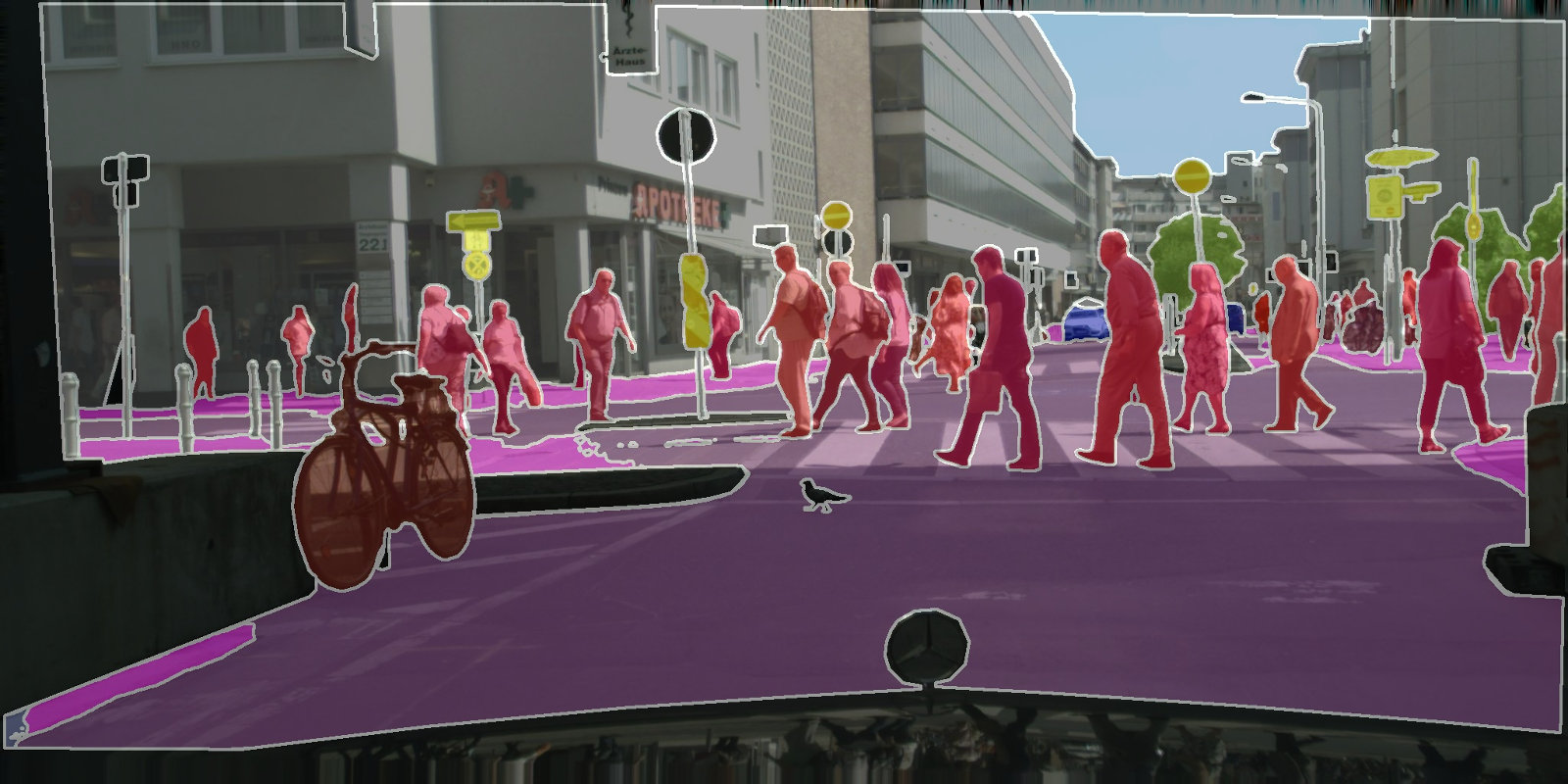} &
    \includegraphics[width=0.33\linewidth]{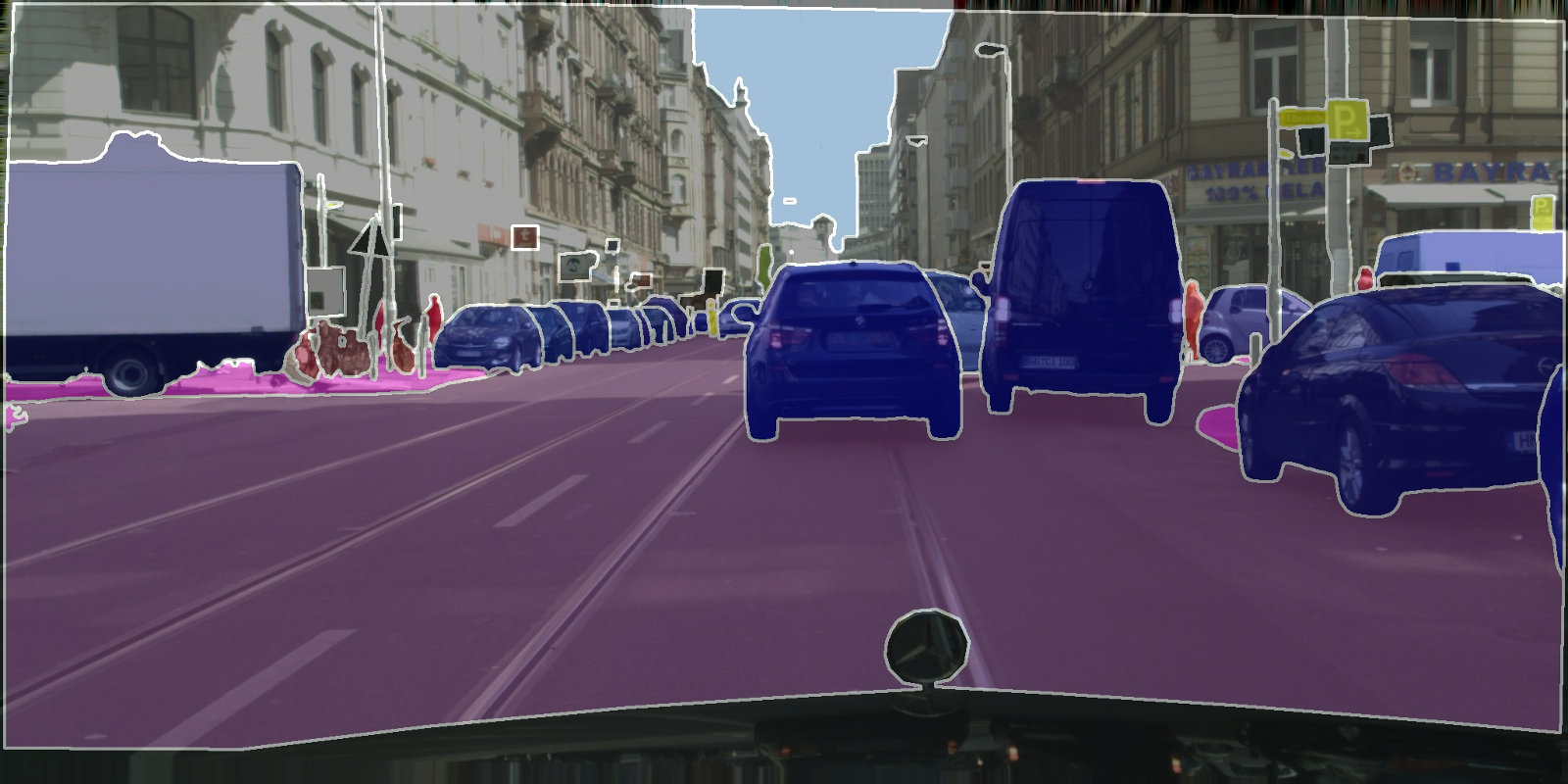} \\
    (a) & (b) & (c) \\
  \end{tabular}
  \caption{Examples of panoptic segmentation results with AdaptIS on Cityscapes dataset. (a) --- notice, how AdaptIS is able to accurately segment objects of complex shape (heatmaps show the output of AdaptIS for the car and the bicycle objects); (b) --- see how AdaptIS can handle severe occlusions in crowded environments; (c) --- though the proposed architecture is built on top of single-scale features, it can segment objects of different sizes.}
  \label{fig:AdaptIS_properties}
\end{figure*}
\endgroup

At test time AdaptIS outputs a pixelwise mask for a single object. However, the desired result is to segment all objects in the image rather than just one of them. For this purpose we provide different point proposals and obtain masks for multiple objects one by one. Since the objects are processed independently, their masks can overlap. To resolve these ambiguities, we use a greedy algorithm described below.

\textbf{Aggregating the results.}
Let $S$ denote a map of segmented objects. In the beginning all elements in $S$ are initialized as ''unknown''. The algorithm proceeds in iterations. At each iteration we sample a point proposal $(x_i, y_i)$, compute a real-valued output of AdaptIS $C_i$, and threshold it to obtain an object mask $M_i = C_i > T$. At the first iteration we simply add an object on the map, \ie mark pixels in $S$ corresponding to $M_1$ as ''segmented''. At the next iterations we compute the intersection of $M_i$ with ''segmented'' area  in $S$. In case the intersection is lower than 50\%, we add the new object on the map. Otherwise we simply ignore the result and move on to the next point proposal. 

The algorithm terminates either when all pixels in $S$ are marked as ''segmented'' or when we run out of point proposals. In the end, each pixel is assigned an instance id that has the highest confidence among all segmented objects: $R(x,y) = \argmaxC_{i=1, \dots, N}C_i(x,y)$. Figure \ref{fig:toy_instance_segmentation} illustrates iterations of the method.

It should be noted that in the described method only a light-weight AdaptIS head needs to be run several times per image, while the backbone can be run only once. This significantly reduces the amount of computations and makes the described method quite applicable in practice.

\textbf{Proposal generation.}
One can notice that if a point proposal $(x_i, y_i)$ is marked as ''segmented'' in $S$, then most likely it will produce an object mask $M_i: (x_i, y_i) \in M_i$ overlapping with an already segmented object. This means that all the points marked as ''segmented'' in $S$ would rather not be selected as point proposals. Therefore, at each iteration of the method described above we sample a new point proposal from the set of ''unknown'' points. 

We explore different strategies for prioritizing the choice of a point proposal at each iteration. For the toy problem in Section \ref{sec:toy} we experiment with simple random sampling, and for panoptic segmentation in Section \ref{sec:panoptic} we introduce a specialized network branch for prioritizing point proposals.

\section{Panoptic segmentation}
\label{sec:panoptic}

\begin{figure}[t]
    \includegraphics[width=\linewidth]{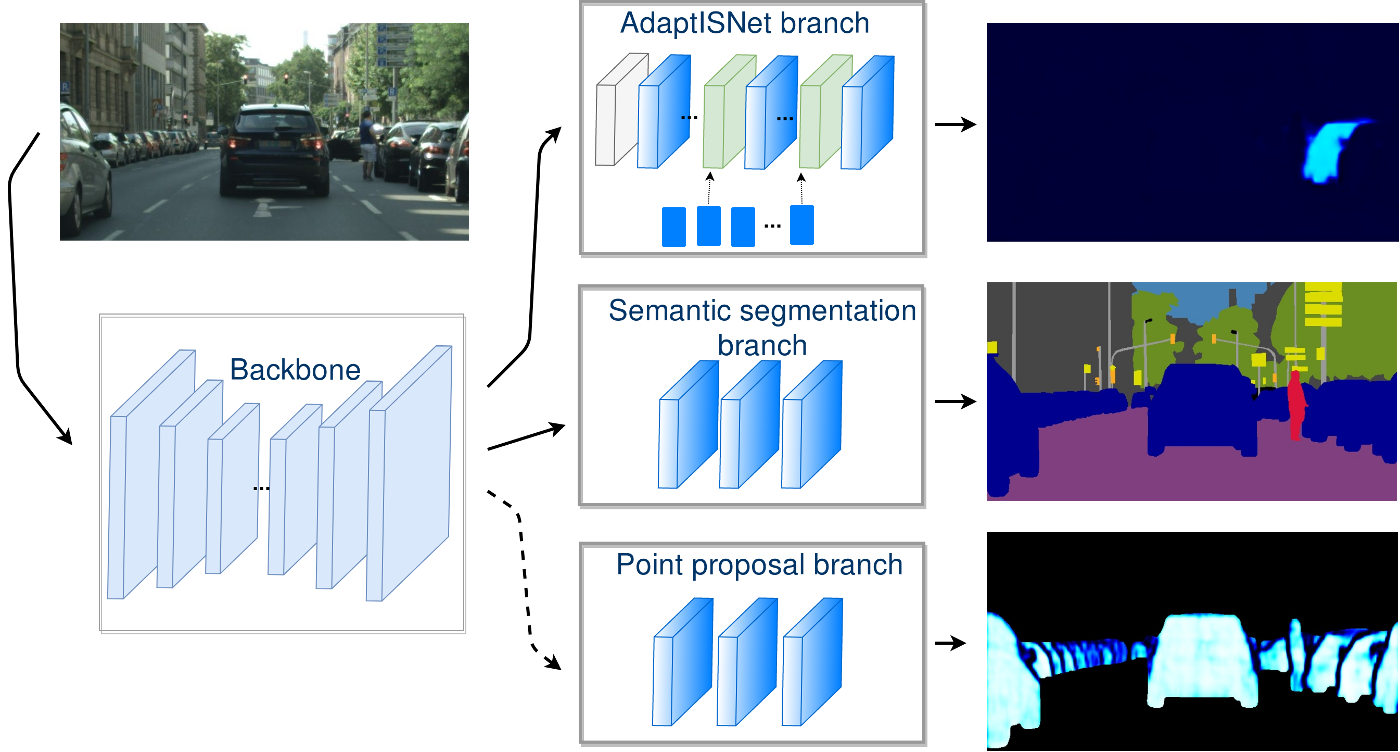}
    \caption{Panoptic segmentation with AdaptIS. The AdaptIS branch for class-agnostic instance segmentation and a standard semantic segmentation branch (\eg PSPNet or DeepLab) are trained jointly with the backbone. The \emph{Point Proposal} branch uses the freezed backbone for training. }
    \label{fig:panoptic_network}
\end{figure}

\textbf{Semantic segmentation branch.} 
AdaptIS does not use semantic labels of the objects for training. However in practice we need to predict object class along with an object mask. For this we pair AdaptIS with standard semantic segmentation pipeline. In this work we train a \emph{Semantic Segmentation Branch} jointly with AdaptIS on a shared backbone (see Figure \ref{fig:panoptic_network}). Below we explain how we use the results of semantic segmentation at test time.

\textbf{Point proposal branch.}
We train a special \emph{Point Proposal Branch}, that predicts the priority of sampling different point proposals. We formulate this task as binary segmentation: the network predicts whether a point $(x, y)$ would make a good or a bad proposal. Point proposal branch has exactly the same structure as the semantic segmentation branch. We train it after the others with a frozen backbone. 

During training we provide ground truth for the point proposal branch as follows. We pick a random object and sample several random point proposals in this object. Then we run AdaptIS with these proposals and compare the results to the ground truth object mask. After that we sort the point proposals by IoU with ground truth mask. The points that fall into top 20\% are marked as positive examples and the others are marked as negative examples. We train the point proposal branch with a simple binary cross-entropy loss function.

\textbf{Obtaining panoptic segmentation result.} 
Panoptic segmentation aims at assigning class label to each image pixel, and an instance id to each pixel classified as a "thing" class \cite{kirillov2018panoptic}. 

Notice that all "stuff" labels can be inferred in one pass of the semantic segmentation branch. Thus, we first run the semantic segmentation branch and obtain the map of all "stuff" classes. Then we proceed with the method for instance segmentation described in Section \ref{sec:inference}. The only difference is that instead of starting with an empty map $S$ we mark all the "stuff" pixels in $S$ as ''segmented''. 

We use the output of the Point Proposal Branch as follows. We are interested in point proposals that get a high score according to the point proposal branch. Point proposal branch provides a dense output of the same size as input image. To reduce the number of proposals for evaluation we first find local maximums in the result of point proposal branch (see Figure \ref{fig:proposals_maximums}). For that we use a standard method based on breadth-first search. In most cases, it finds about a hundred local maximums per image, which we use as point proposals. At each iteration of the method from Section \ref{sec:inference} we remove the point proposals that have been marked as ''segmented'' at the previous steps.

After assigning instance ids, we need to obtain semantic labels for each instance. Each pixel is assigned a class label $l \in \{1,...,K\}$ that has the highest average confidence among all labels for a mask $M$: 
$l(M) = \argmaxC_{l=1,...,K}\sum_{(x, y) \in M}p_{x, y}^l$,
where $p_{x, y}^l$ is the softmax output of the semantic segmentation branch for the $l$-th class at pixel $(x, y)$.

\begin{figure}[t]
    \includegraphics[width=\linewidth]{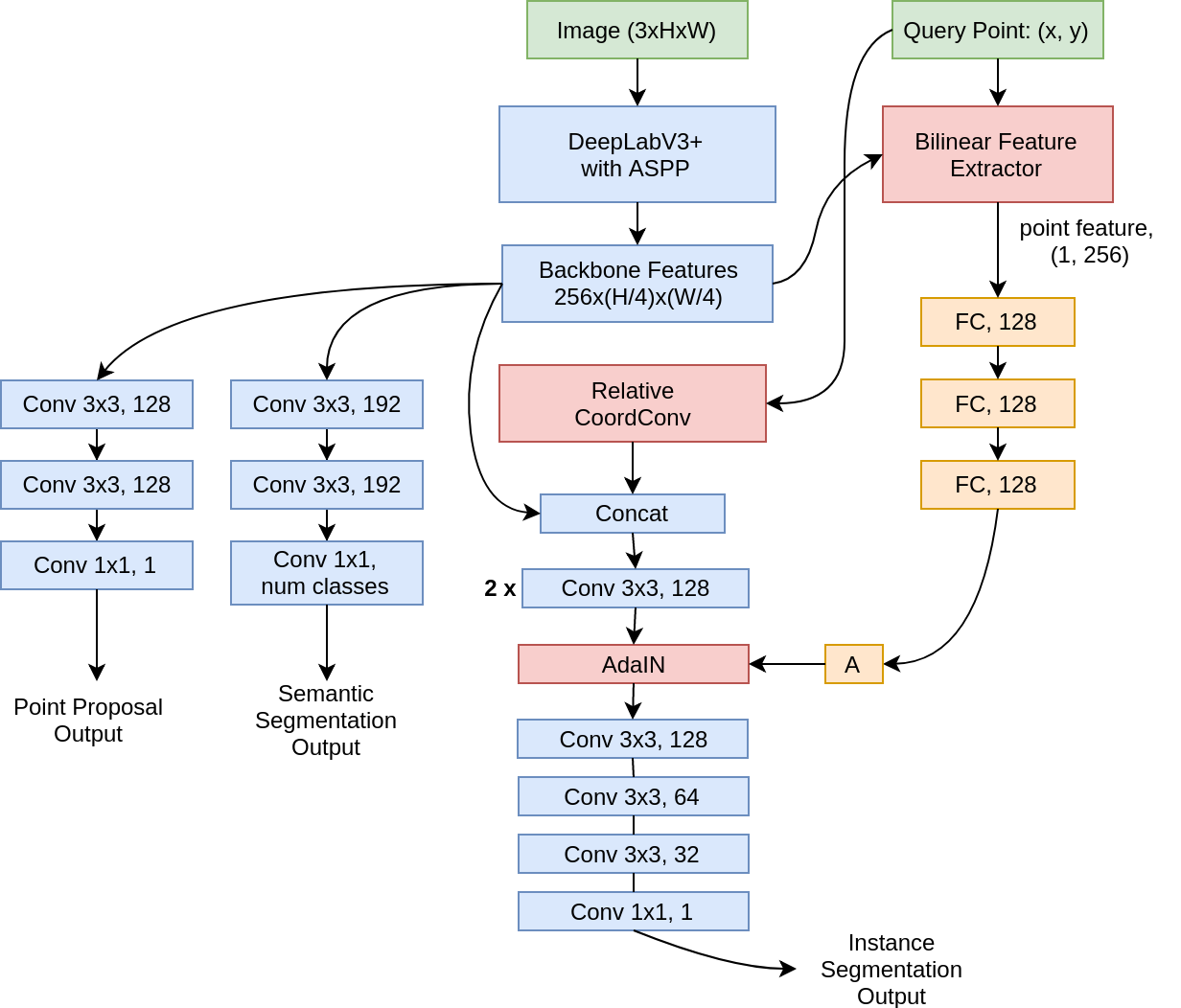}
    \caption{Architecture of AdaptIS for panoptic segmentation that we used in our experiments for ResNet-50 backbone. Each conv layer is followed by ReLU activation and BatchNorm. }
    \label{fig:network_diagram}
\end{figure}

\section{Experiments on toy problem}
\label{sec:toy}

\textbf{Toy data.}
To showcase the limitations of detection-first methods we have designed a challenging toy problem. We want to generate many overlapping objects that have very similar bounding boxes. For that we generate images of size 96x96 pixels containing from 8 to 22 elongated objects that resemble bacteria or biological cells. All of those objects are of the same color (blue with red boundary) and slightly varied sizes. The position and orientation of those objects are sampled at random. To make the problem more difficult we added random blur and random high frequency noise to the images. Train set contains 10000 images, and 2000 are left for testing.

\begin{table}
\small
\begin{center}
\begin{tabular}{l|c|c|c|c}
\hline
 & AP@.5 & AP@.7 & AP@.8 & AP@.9 \\
\hline
\hline
Mask R-CNN  &  74.6\%  &  59.4\%  &  43.8\%  &  5.8\%  \\
AdaptIS  &  \textbf{99.6\%}  &  \textbf{99.2\%}  &  \textbf{98.8\%}  &  \textbf{97.5\%}  \\
\hline
\end{tabular}
\end{center}
\caption{Comparison of AdaptIS with Mask R-CNN on toy problem. Toy images contain multiple overlapping objects with coinciding bounding boxes, and hence are difficult for detection-first methods. At the same time, AdaptIS can segment those objects almost perfectly well.}
\label{tab:toy_comparison}
\end{table}

\textbf{Evaluation protocol and results.}
In this experiment we compared AdaptIS with U-Net backbone to Mask R-CNN with ResNet-50 ImageNet pre-trained backbone. For Mask R-CNN we upsample an image by a factor of 4, \ie to 384x384. \mbox{AdaptIS} is trained and runs on original images. Mask R-CNN was trained for 55 epoch starting with SGD with learning rate of 0.001 and dropping it by factor of 10 at 35 and 45 epochs. We train AdaptIS for 140 epochs from scratch with Adam  with learning rate of 0.0005, $\beta_1 = 0.9$, $\beta_2 = 0.999$. 

For the method described in Section \ref{sec:inference} we used simple random sampling of point proposals. At each iteration of the method we sampled 7 random points and ran AdaptIS for each of them. Then among the 7 results we chose the one with the highest average confidence of the object mask.

We measure average precision (AP) at different IoU thresholds. Table \ref{tab:toy_comparison} shows results of AdaptIS and Mask R-CNN. The main source of errors of Mask R-CNN is heavy overlap of bounding boxes. In our toy dataset objects often have very similar bounding boxes, hence Mask R-CNN has no way to distinguish these objects. At the same time AdaptIS can deal with severe occlusions because it is provided with query points belonging to these objects, which allows it to segment them. We notice that AP drastically drops for Mask R-CNN at high IoU thresholds that indicates inaccurate masks shapes due to reduction in feature dimensionality in ROI Pooling.
See examples of the images and results of AdaptIS and Mask R-CNN in Figure \ref{fig:toy_examples}.

\begingroup
\setlength{\tabcolsep}{0.8pt} 
\renewcommand{\arraystretch}{0.5} 
\begin{figure}[t]
 \begin{tabular}{cc}
    \includegraphics[width=0.48\linewidth]{images/proposals_maximums/point_proposals_image.jpg} &
    \includegraphics[width=0.48\linewidth]{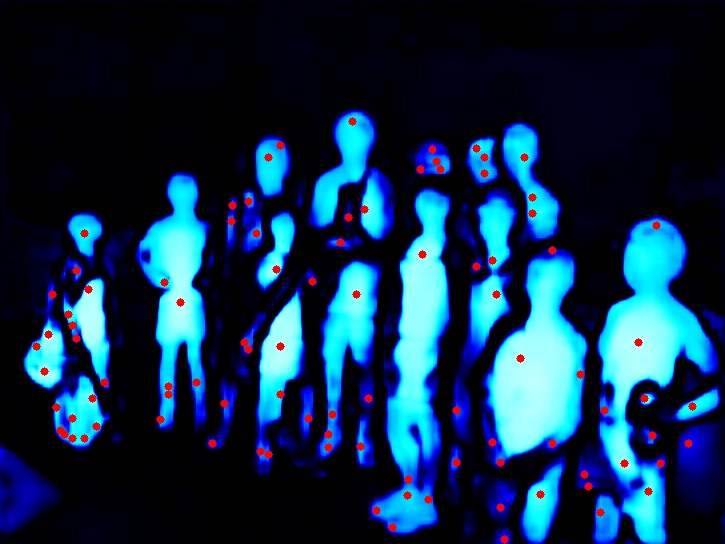} \\
    (a) & (b) \\
 \end{tabular} 
 \caption{Point proposals generation: (a) --- input image; (b) --- output of the point proposal branch with detected local maximums.}   
 \label{fig:proposals_maximums}
\end{figure}
\endgroup

\begingroup
\setlength{\tabcolsep}{1pt} 
\begin{figure*}[h]
 \begin{tabular}{cccc}
    \includegraphics[width=0.23\linewidth]{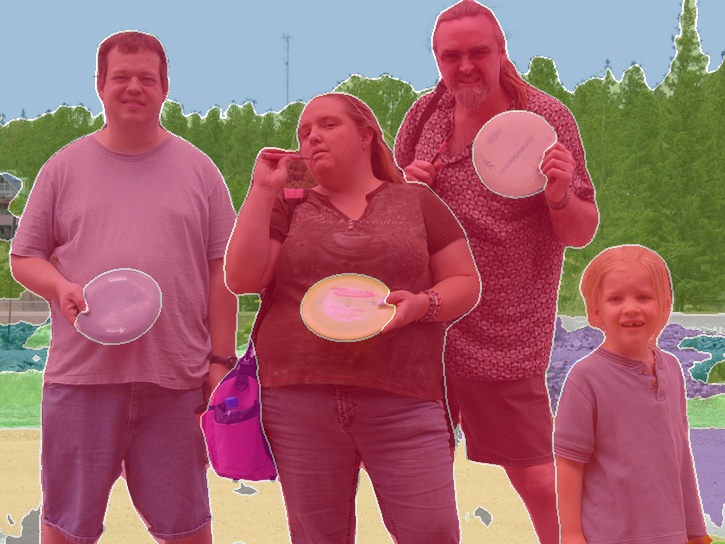} &
    \includegraphics[width=0.26\linewidth]{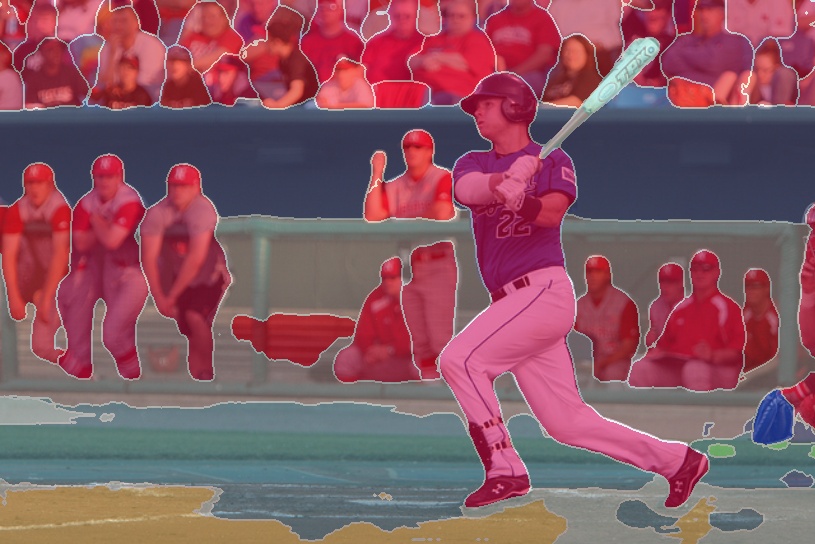} &
    \includegraphics[width=0.26\linewidth]{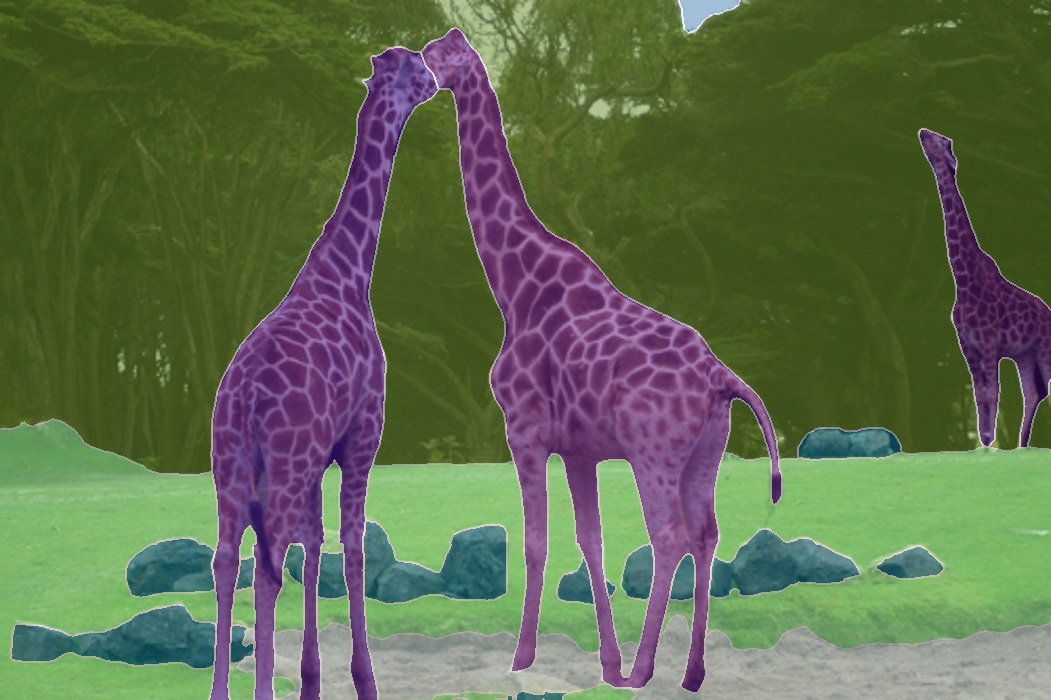} &
    \includegraphics[width=0.23\linewidth]{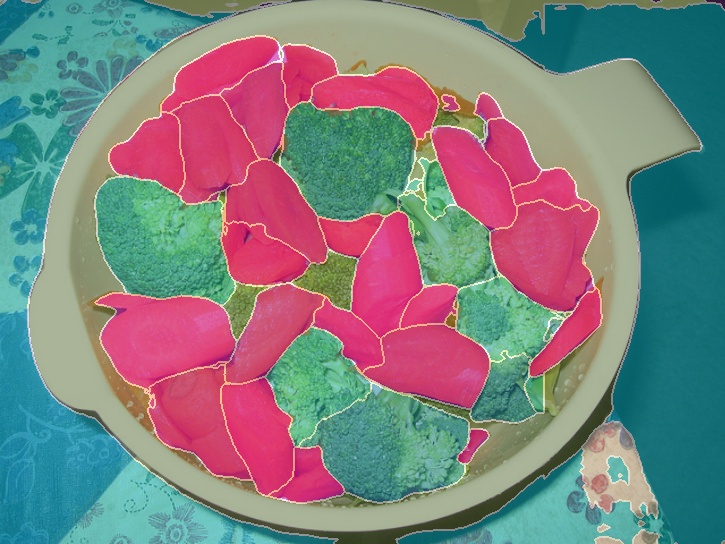} \\
  \end{tabular}
  \caption{Examples of panoptic segmentation results on COCO dataset.}
  \label{fig:coco_examples}
  
 \begin{tabular}{cccc}
    \includegraphics[width=0.30\linewidth]{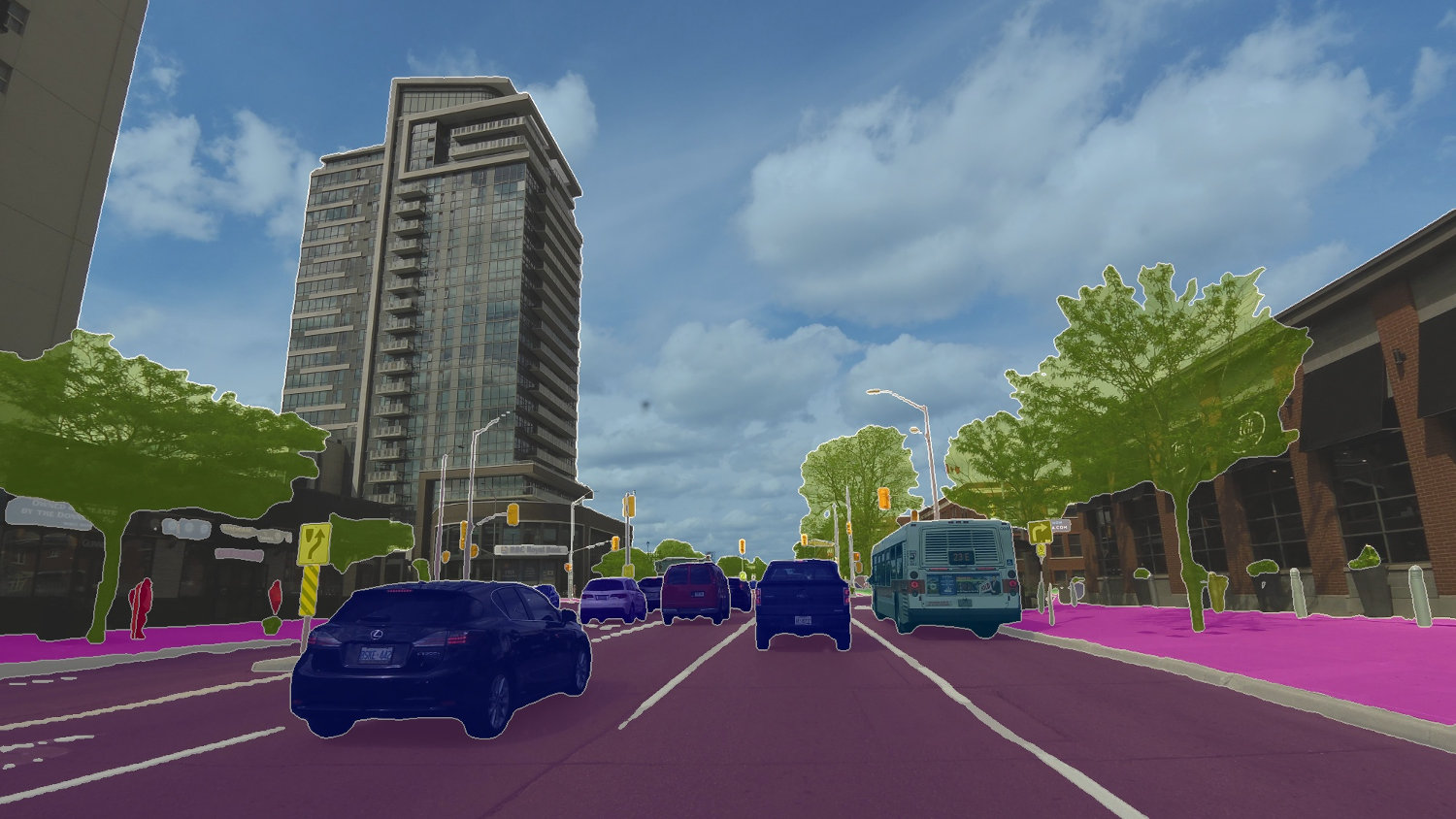} &
    \includegraphics[width=0.225\linewidth]{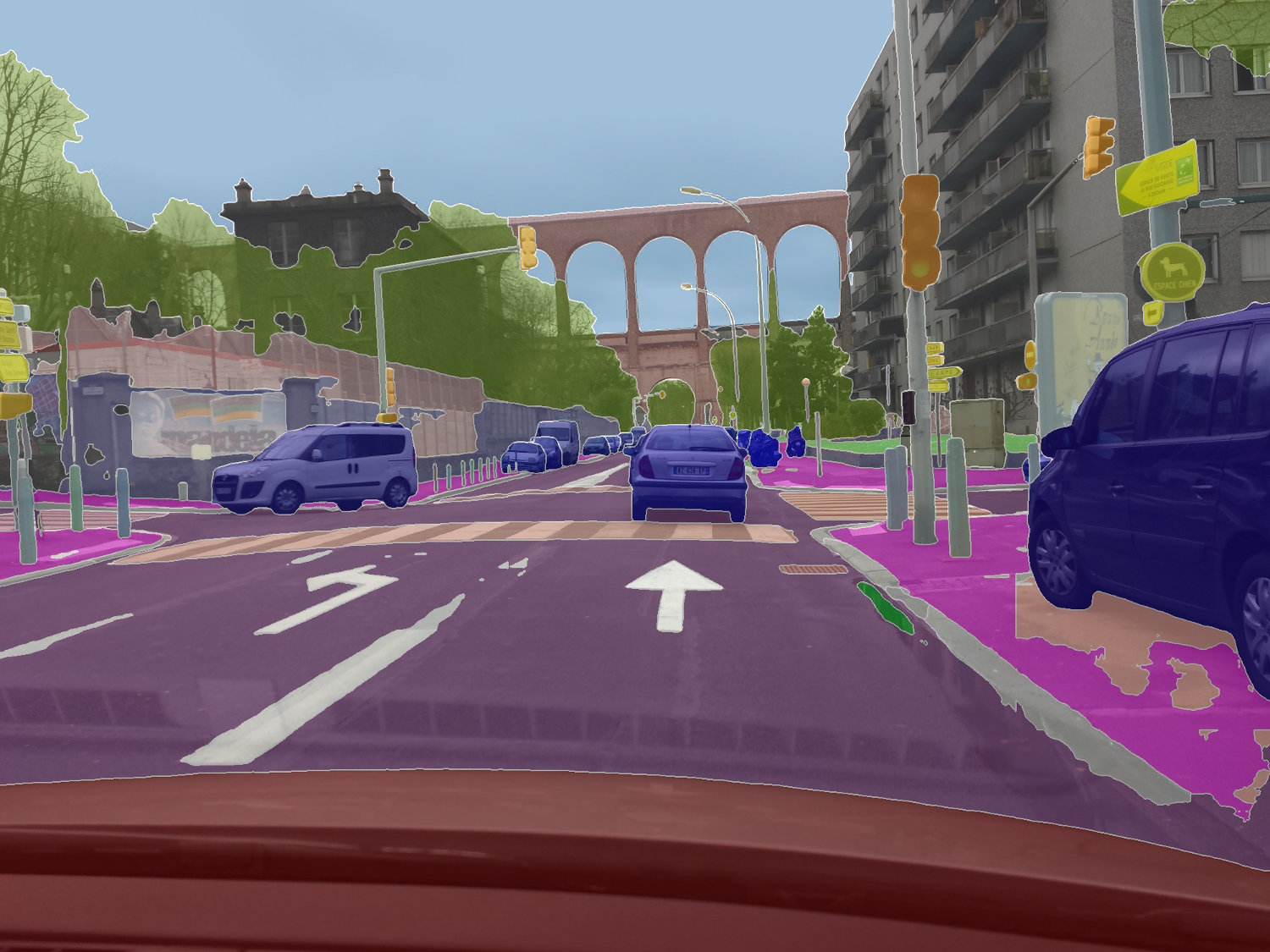} &
    \includegraphics[width=0.225\linewidth]{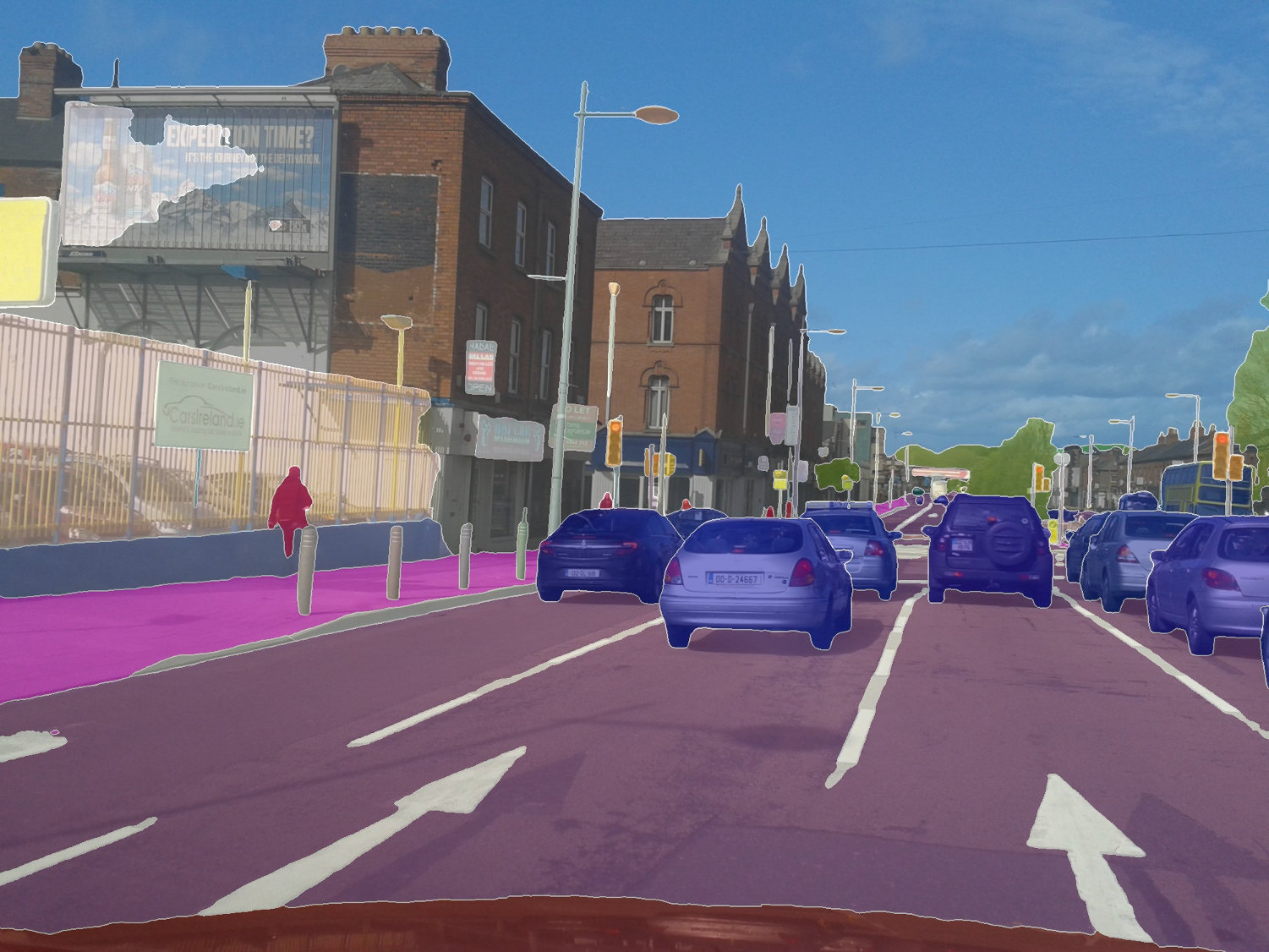} &
    \includegraphics[width=0.225\linewidth]{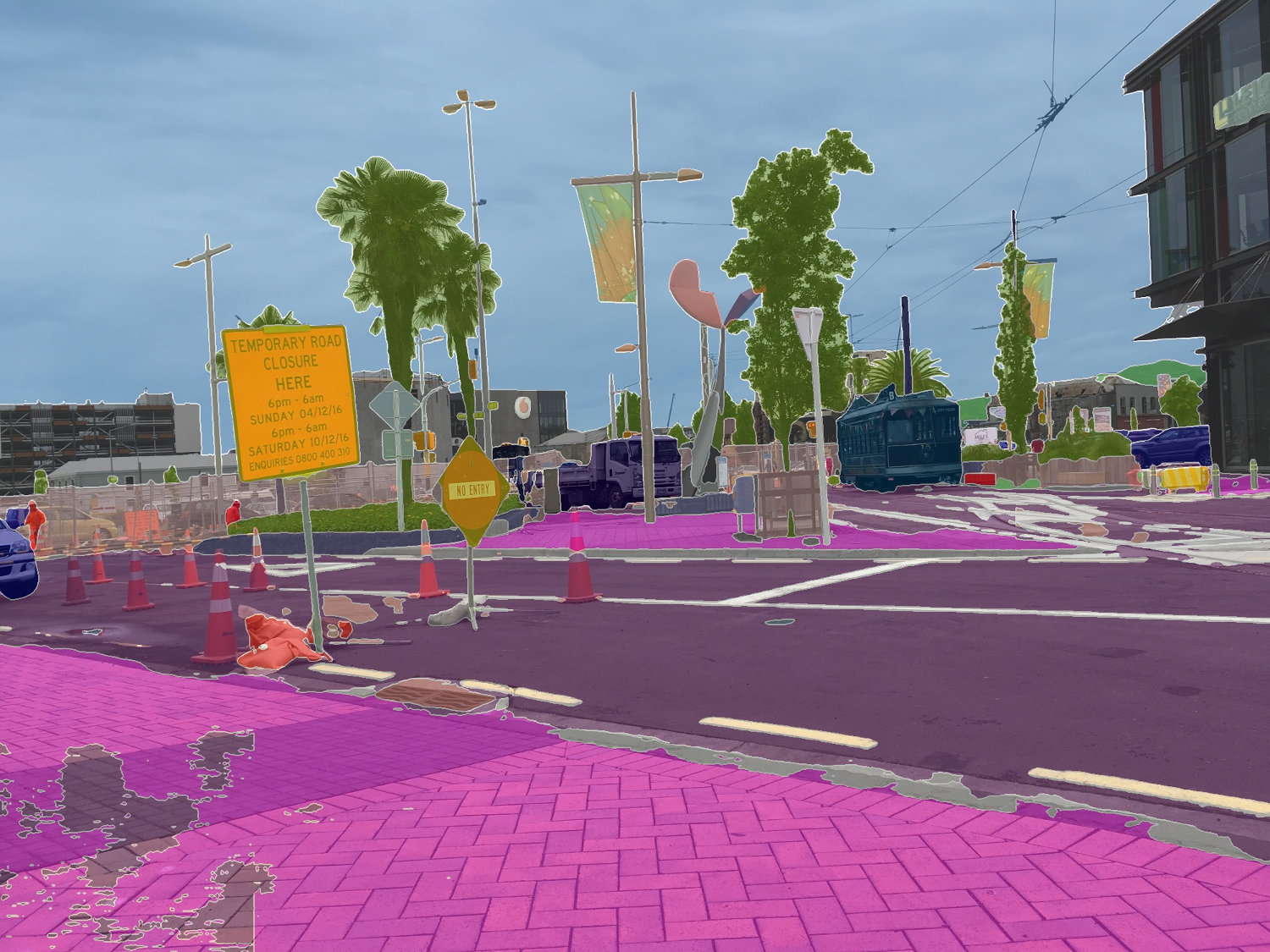} \\
  \end{tabular}
  \caption{Examples of panoptic segmentation results on Mapillary dataset.}
  \label{fig:mapillary_examples}
\end{figure*}
\endgroup

\begin{table*}[t]
\small
\begin{center}
\begin{tabular}{l|c|c|c|c|c|c|c|c}
\hline
Method & Pre-training & Backbone & \makecell{m/s \\ test} & $\text{PQ}$  & $\text{PQ}^{st}$ & $\text{PQ}^{th}$ & $\text{mIoU}$ & $\text{AP}$ \\
\hline
\hline
Mask R-CNN \cite{he2017mask} & ImageNet & ResNet-50 &  & - & - & - & - & 31.5 \\
Mask R-CNN \cite{he2017mask} & ImageNet+COCO & ResNet-50 &  & - & - & - & - & \emph{36.4} \\
\hline
TASCNet \cite{li2018learning} & ImageNet  & ResNet-50 &  & 55.9 & 59.8 & 50.6 & - & - \\
UPSNet \cite{xiong2019upsnet} & ImageNet & ResNet-50 &  & 59.3 & 62.7 & 54.6 & 75.2 & 33.3 \\
CRF + PSPNet \cite{li2018weakly} & ImageNet & ResNet-101 &  & 53.8 & 62.1 & 42.5 & 71.6 & - \\
Panoptic FPN \cite{kirillov2019panoptic} & ImageNet & ResNet-101 &  & 58.1 & 62.5 & 52.0 & 75.7 & 33.0 \\
DeeperLab \cite{yang2019deeperlab} & ImageNet & Xception-71 &  & 56.5 & - & - & - & - \\
TASCNet \cite{li2018learning} & ImageNet+COCO  & ResNet-50 &  & \emph{59.2} & \emph{61.5} & \emph{56.0} & \emph{77.8} & \emph{37.6} \\
TASCNet-multiscale \cite{li2018learning} & ImageNet+COCO & ResNet-50 & + & \emph{60.4} & \emph{63.3} & \emph{56.1} & \emph{78.0} & \emph{39.0} \\
MRCNN + PSPNet \cite{kirillov2018panoptic} & ImageNet+COCO & ResNet-101 &  & \emph{61.2} & \emph{66.4} & \emph{54} & - & \emph{36.4}  \\
UPSNet \cite{xiong2019upsnet} & ImageNet+COCO & ResNet-101 &  & \emph{60.5} & \emph{63.0} & \emph{57.0} & \emph{77.8} & \emph{37.8} \\
UPSNet-multiscale \cite{xiong2019upsnet} & ImageNet+COCO & ResNet-101 & + & \emph{61.8} & \emph{64.8} & \emph{57.6} & \emph{79.2} & \emph{39.0} \\
\hline
AdaptIS (ours) & ImageNet  & ResNet-50 &  & 59.0  &  61.3  &  55.8  & 75.3   & 32.3  \\ 
AdaptIS (ours) & ImageNet & ResNet-101 &  & 60.6  &  62.9  &  57.5  & 77.2 & 33.9 \\ 
AdaptIS (ours) & ImageNet & ResNeXt-101 &  & \textbf{62.0}  &  \textbf{64.4}  &  \textbf{58.7}  & \textbf{79.2}  &  \textbf{36.3} \\
\hline
\end{tabular}
\end{center}
\caption{Evaluation results on Cityscapes \emph{val} dataset.}
\label{tab:evaluation_cityscapes}
\end{table*}

\section{Experiments on standard benchmarks}

We evaluate our panoptic segmentation pipeline on the standard panoptic segmentation benchmarks: Cityscapes \cite{cordts2016cityscapes}, Mapillary \cite{neuhold2017mapillary} and COCO \cite{lin2014microsoft}. We present results on validation sets for all of those benchmarks and compare to state-of-the-art. We compute standard metrics for panoptic segmentation (namely, $\text{PQ}$ --- panoptic quality, $\text{PQ}^{st}$ -- PQ stuff, $\text{PQ}^{th}$ --- PQ things). For Cityscapes following the other works we additionally report the $\text{mIoU}$ and $\text{AP}$, which are standard for this dataset. We use DeepLabV3+ \cite{chen2017rethinking} architecture as backbone for all experiments, see more details in Figure \ref{fig:network_diagram}.

\begin{table}[t]
\small
\begin{center}
\begin{tabular}{c|c|c|c|c}
\hline
Method & Backbone & $\text{PQ}$ & $\text{PQ}^{st}$ & $\text{PQ}^{th}$ \\
\hline
\hline
\cite{de2018panoptic} & ResNet-50 & 17.6 & 27.5 & 10.0 \\
\cite{yang2019deeperlab} & Xception-71 & 31.9 & - & - \\
\cite{li2018learning} & ResNet-50 & \emph{30.7} & \emph{33.4} & \emph{28.6} \\
\hline
Ours & ResNet-50 & 32.0 & 39.1 & 26.6 \\
Ours & ResNet-101 & 33.4 & 40.2 & 28.3 \\
Ours & ResNeXt-101 & \textbf{35.9} & \textbf{41.9} & \textbf{31.5} \\
\hline
\end{tabular}
\end{center}
\caption{Evaluation results on Mapillary \emph{val} dataset.}
\label{tab:evaluation_mapillary}
\end{table}

\begin{table}[t]
\small
\begin{center}
\begin{tabular}{c|c|c|c}
\hline
Method & $\text{PQ}$ & $\text{PQ}^{st}$ & $\text{PQ}^{th}$ \\
\hline
\hline
liuxu & \textbf{41.1} & \textbf{45.6} & 37.8 \\
jie.li & 38.6 & 38.2 & \textbf{38.9} \\
\hline
Ours & 36.8 & 41.4 & 33.3 \\
\hline
\end{tabular}
\end{center}
\caption{Evaluation results on Mapillary \emph{test-dev} dataset. We show the first 3 rows in the leaderboard.}
\label{tab:evaluation_mapillary_test}
\end{table}

\textbf{Cityscapes} dataset has 5000 images of ego-centric driving scenarios in urban settings which are split into 2975, 500 and 1525 images for training, validation and testing respectively. It consists of 8 and 11 classes for thing and stuff. In all our experiments the panoptic segmentation network was trained on \emph{train} part and evaluated on \emph{val} part. All images have the same resolution $1024\times2048$.

\textbf{Mapillary} dataset is a street scene dataset like Cityscapes. It consists of a wide variety of geographic settings, camera types, weather conditions, image aspect ratios, and object frequencies. The average resolution of the images is approximately 9 megapixels, varying from $1024\times768$ to higher than $4000\times6000$. It contains 65 semantic classes, including 37 thing classes and 28 stuff classes. We train on the \emph{train} part of the dataset containing 18000 training images and evaluate the method on \emph{val} part containing 2000 validation images. 

\textbf{COCO} dataset for panoptic segmentation task consists of 80 and 53 classes for thing and stuff respectively.  We train on approximately 118k images from \emph{train2017} and present results on 5k images from \emph{val2017}. 

\textbf{Cityscapes training.} We train AdaptIS and semantic segmentation branches for 260 epochs and dropping learning rate by factor of 10 at 240 and 255 epochs. For ResNet-50 training we use 2 GPUs with batch size 8, for ResNet-101 and ResNeXt-101 \cite{xie2017aggregated} we use 4 GPUs with batch size 8. We sample 6 point proposals per image during training. We train networks on random crops of size of $400\times800$. We use scale jitter augmentation with scale factor varying from 0.2 to 1.2. Also we use random contrast, brightness and color augmentation. Point proposal network was trained for 10 epochs. 

\textbf{Mapillary training.} We train only ResNeXt-101 model for this dataset on 8 GPUs with batch size 8 for 65 epochs and dropping learning rate by factor of 10 at 50 and 60 epochs. We scale input images by largest side varied from 500 to 2200 pixels and then perform random crop of size $400\times800$. The dataset is large enough, therefore we use only horizontal flip augmentation. We sample 6 point proposals per image during training.

\textbf{COCO training.}  We train only ResNeXt-101 model for this dataset on 8 GPUs with batch size 16 for 20 epochs and dropping learning rate by factor of 10 at 15 and 19 epochs. We scale input images by shortest side varied from 350 to 700 pixels and then perform random crop of size $544\times544$. We use horizontal flip augmentation. We sample 8 point proposals per image during training.

\textbf{Inference details.} For Cityscapes we use images of original resolution. For Mapillary we downscale all images to 2000 pixels by longest side. For COCO we rescale images to 600 pixels by shortest side. The greedy algorithm from \mbox{Section \ref{sec:inference}} is applied with threshold $T=0.40$ for Cityscapes and $T=0.35$ for COCO and Mapillary. We use only horizontal flip augmentation at test time and do not use multiscale testing for all datasets.

\textbf{Implementation details.} We use MXNet Gluon \cite{chen2015mxnet} with GluonCV \cite{zhang2019bag} framework for training and inference of our models. In all our experiments we use Adam  optimizer with the starting learning rate $5 \times 10^{-5}$ for pre-trained backbone and $10^{-4}$ for all other parameters, \mbox{$\beta_1 = 0.9$}, \mbox{$\beta_2 = 0.999$}. Semantic segmentation and AdaptIS branches are trained jointly with the backbone. The point proposal network was trained afterwards with frozen backbone, semantic segmentation and AdaptIS branches. We sample 48 point proposals per object during training for all datasets. For all experiments we set the radius of Relative CoordConv to 280 pixels.

\textbf{Results and discussion.} \mbox{Table \ref{tab:evaluation_cityscapes}} presents comparison of AdaptIS to recent works on panoptic segmentation on Cityscapes dataset. We also add Mask R-CNN for comparison as it is the basis of many works on panoptic segmentation. Interestingly, AdaptIS shows state-of-the-art accuracy in terms of PQ even without pre-training on COCO and multiscale testing. Though we did not adapt the method for Mapillary and COCO datasets, we improve upon state-of-the-art by more than 3\% in terms of PQ on Mapillary dataset (see \mbox{Table \ref{tab:evaluation_mapillary}}) and show metrics close to the results of very recent UPSNet (see \mbox{Table \ref{tab:evaluation_mscoco}}). The results on \mbox{\emph{test-dev}} part of the Mapillary and COCO datasets are shown in \mbox{Table \ref{tab:evaluation_mscoco_test}} and \mbox{Table \ref{tab:evaluation_mapillary_test}}.

\begingroup
\setlength{\tabcolsep}{3pt} 
\begin{table}[t]
\small
\begin{center}
\begin{tabular}{l|c|c|c|c|c}
\hline
Method & Backbone & \makecell{m/s \\ test} & $\text{PQ}$ & $\text{PQ}^{st}$ & $\text{PQ}^{th}$ \\
\hline
\hline
JSIS-Net \cite{de2018panoptic} & ResNet-50 &  & 26.9 & 23.3 & 29.3 \\
AUNet \cite{li2018attention} & ResNet-50 &  & 39.6 & 25.2 & \textbf{49.1} \\
DeeperLab \cite{yang2019deeperlab} & Xception-71 &  & 33.8 & - & - \\
UPSNet \cite{xiong2019upsnet} & ResNet-101 &  & \textbf{42.5} & \textbf{33.4} & 48.5 \\
UPSNet-ms \cite{xiong2019upsnet} & ResNet-101 & + & \emph{43.2} & \emph{34.1} & \emph{49.1}\\
\hline
AdaptIS & ResNet-50 & &  35.9	& 29.3 & 40.3 \\
AdaptIS & ResNet-101 & & 37.0 & 29.9 & 41.8 \\
AdaptIS & ResNeXt-101 & & 42.3 & 31.8 & 49.2 \\
\hline
\end{tabular}
\end{center}
\caption{Evaluation results on COCO \emph{val} dataset.}
\label{tab:evaluation_mscoco}
\end{table}
\endgroup

\begingroup
\setlength{\tabcolsep}{3pt} 
\begin{table}[t]
\small
\begin{center}
\begin{tabular}{l|c|c|c|c}
\hline
Method & Backbone & $\text{PQ}$ & $\text{PQ}^{st}$ & $\text{PQ}^{th}$ \\
\hline
\hline
JSIS-Net \cite{de2018panoptic} & ResNet-50 & 27.2 & 23.4 & 29.6 \\
AUNet \cite{li2018attention} & ResNeXt-152-FPN & 46.5 & 32.5 & 55.9 \\
UPSNet-ms \cite{xiong2019upsnet} & ResNet-101-FPN & \textbf{46.6} & \textbf{36.7} & \textbf{53.2} \\
\hline
AdaptIS & ResNeXt-101 & 42.8 & 31.8 & 50.1 \\
\hline
\end{tabular}
\end{center}
\caption{Evaluation results on COCO \emph{test-dev} dataset.}
\label{tab:evaluation_mscoco_test}
\end{table}
\endgroup

\section{Conclusion and future work}
We introduce a novel network architecture for instance and panoptic segmentation that is conceptually different from mainstream detection-first instance segmentation methods. Presented architecture in many ways overcomes the limitations of existing methods and shows good performance on standard panoptic segmentation benchmarks. We hope that this work could serve as a strong foundation for future research.

\addtocontents{toc}{\protect\contentsline{section}{Appendix:}{}}
\begin{appendices}

\section{Normalized Focal Loss}
\label{appendix_loss}

Given an image and a point $(x, y)$ AdaptIS network produces a mask of an object located at $(x, y)$. We cast this problem as a binary semantic segmentation. The whole network is end-to-end trainable and optimizes just one loss function, that compares network output to the ground truth object mask.

Generally, binary cross entropy (BCE) loss is used for similar problems \cite{yang2019deeperlab}. However, \cite{lin2017focal} showed that BCE pays more attention to pixels that are already correctly classified. \cite{lin2017focal} proposed Focal Loss to overcome this problem. 

The idea of Focal Loss is as follows. Let $M$ denote the ground truth mask of an object, and $\hat{M}$ denote the output of the network. Let us denote the confidence of the network for the correct output at the point $(i, j)$ by $p_{i,j} = p\big(\hat{M}(i, j) = M(i, j)\big)$. The original Focal Loss at the point $(i, j)$ is defined as
$$
    \text{FL}(i, j) = -(1-p_{i, j})^\gamma \log p_{i, j}.
$$

Consider the total weight of the values of loss function for all pixels in the image $P(\hat{M}) = \sum_{i, j} (1-p_{i, j})^\gamma$. One can see that $P(\hat{M})$ decreases when accuracy of the network improves. This means that the total gradient of the Focal Loss fades over time. As a consequence the training with Focal Loss is slowed down over training iterations.

In our experiments we use a modification of Focal Loss that we call \emph{Normalized Focal Loss} (NFL). Given an output of the network $\hat{M}$ we define NFL at each pixel $(i, j)$ of the image as follows:
$$
    \text{NFL}(i, j, \hat{M}) = -\dfrac{1}{P(\hat{M})} (1-p_{i, j})^\gamma \log p_{i, j}.
$$

One can notice that the total gradient of the NFL is equal to the total gradient of BCE. Like original Focal Loss, NFL concentrates on the pixels that are misclassified by the network, but  it leads to faster convergence and better accuracy (see Table \ref{tab:ablation_cityscapes} for ablation studies on Cityscapes). 

\section{Ablation studies}

We perform ablation studies to measure the influence of the loss function and the use of Relative CoordConv on the resulting accuracy. We perform experiments on Citiscapes dataset with ResNet-50 backbone. We try the following modifications to the AdaptIS architecture: 1) excluding Relative CoordConv block leads to approximately $5\%$ drop in $PQ$; 2) using simple Focal loss instead of Normalized Focal Loss leads to approximately $3\%$ drop in $PQ$; 3) replacing NFL with cross-entropy leads to minor degradation of PQ mainly due to $PQ^{things}$. Table \ref{tab:ablation_cityscapes} contains the results of theses experiments.

\begin{table}
\begin{center}
\begin{tabular}{l|c|c|c|c}
\hline
Loss & \makecell{Relative \\ CoordConv}&  $\text{PQ}$, \%  & $\text{PQ}^{st}$, \% & $\text{PQ}^{th}$, \% \\
\hline
\hline
NFL & + & \textbf{59.0} & \textbf{61.3} & \textbf{55.8} \\
NFL & - &  53.9  &  61.7  &  43.1  \\
FL & + &  55.9  &  60.1  &  50.1  \\
BCE & + &  58.7  &  61.5  &  54.8  \\
\hline
\end{tabular}
\end{center}
\caption{Ablation studies of panoptic segmentation with AdaptIS on Cityscapes dataset. See text for more details.}
\label{tab:ablation_cityscapes}
\end{table}

\section{Analysis of results on COCO}
 
 For COCO and Mapillary datasets we used the same architecture as for Cityscapes and did not fine-tune any of the parameters. Actually, in the paper we present the results for the first models that we trained. Surprisingly, despite this, we have achieved state-of-the art accuracy on Mapillary \emph{val} and competitive results on COCO \emph{val}. We believe that the results on COCO can be substantially improved by adding more conv layers to the AdaptIS network and fine-tuning the parameters. 
 
 We also notice that most image in COCO contain just one instance of the same object. For the detection-first methods like Mask R-CNN this is not a problem because they learn to segment each object independently. But for AdaptIS one rather needs to provides examples of multiple instances on the same image. Perhaps sampling such images more frequently during training could help. See examples of failure cases on COCO in Figure \ref{fig:coco_failure}.
 
\end{appendices}

\begingroup
\setlength{\tabcolsep}{1pt} 
\renewcommand{\arraystretch}{0.5} 
\begin{figure*}[t]  
 \begin{tabular}{ccc}    
     \includegraphics[width=5.85cm, keepaspectratio]{images/coco_bad_ones/1035_0_image.jpg}  &  
     \includegraphics[width=5.85cm, keepaspectratio]{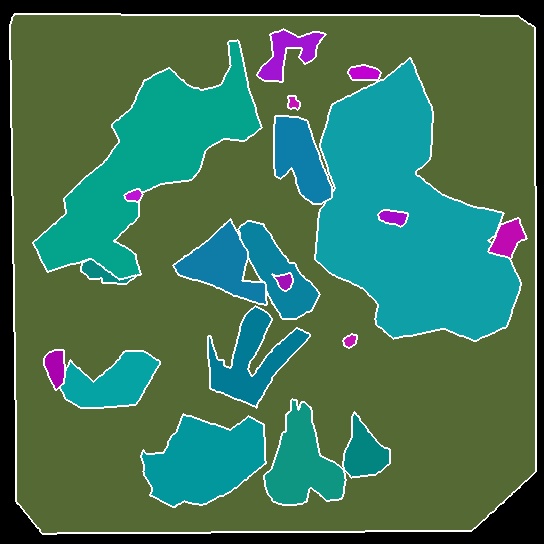}  &    
     \includegraphics[width=5.85cm, keepaspectratio]{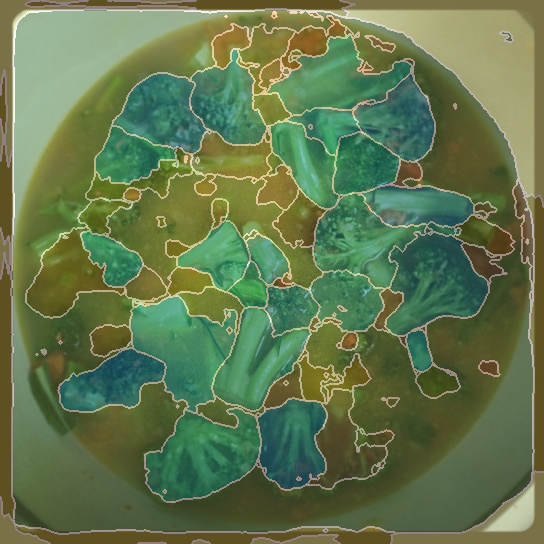}  
 \end{tabular}
 
 \begin{tabular}{ccc}  
     \includegraphics[width=5.85cm, keepaspectratio]{images/coco_bad_ones/1032_0_image.jpg} &   
     \includegraphics[width=5.85cm, keepaspectratio]{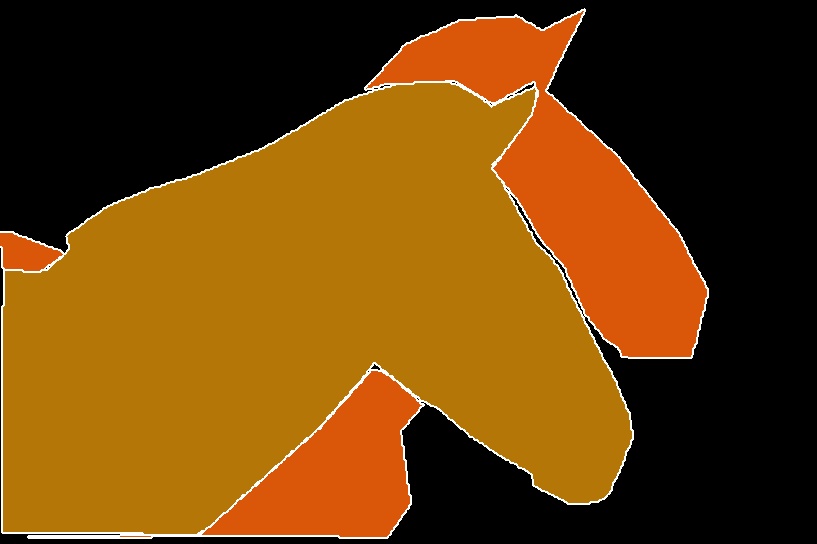} &   
     \includegraphics[width=5.85cm, keepaspectratio]{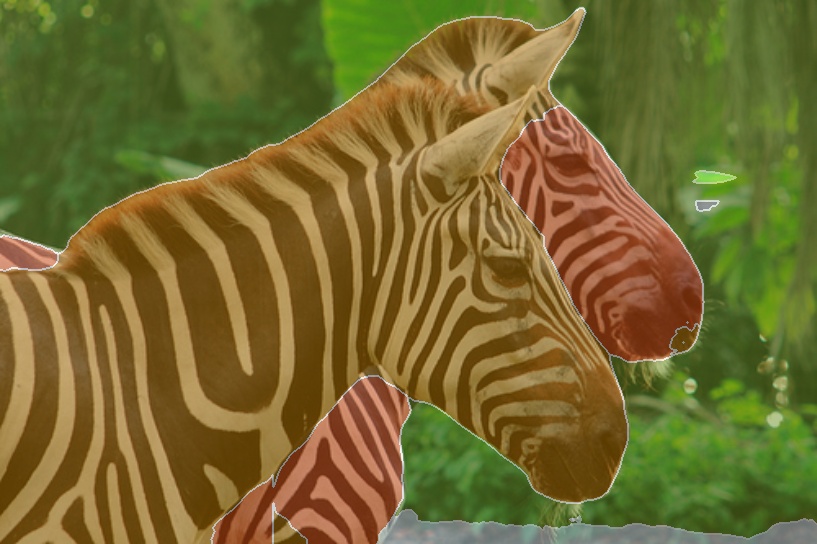} \\
 \end{tabular}
  
 \begin{tabular}{ccc}
    \includegraphics[width=5.85cm, keepaspectratio]{images/coco_bad_ones/1057_0_image.jpg}  &
    \includegraphics[width=5.85cm, keepaspectratio]{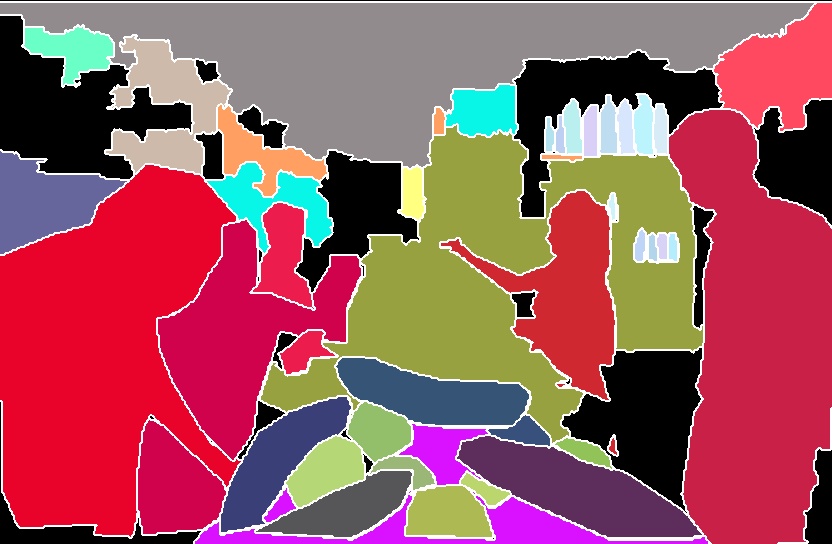}  &
    \includegraphics[width=5.85cm, keepaspectratio]{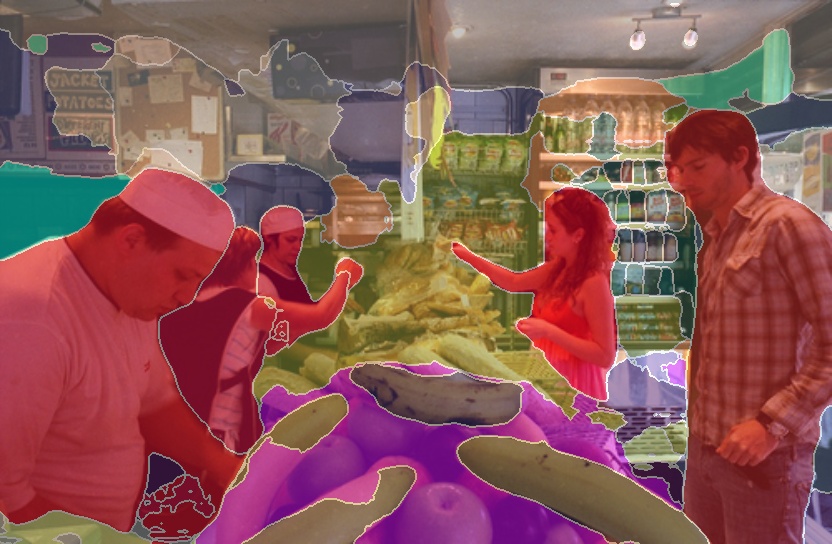}  \\
 \end{tabular}
  
 \begin{tabular}{ccc}
    \includegraphics[width=5.85cm, keepaspectratio]{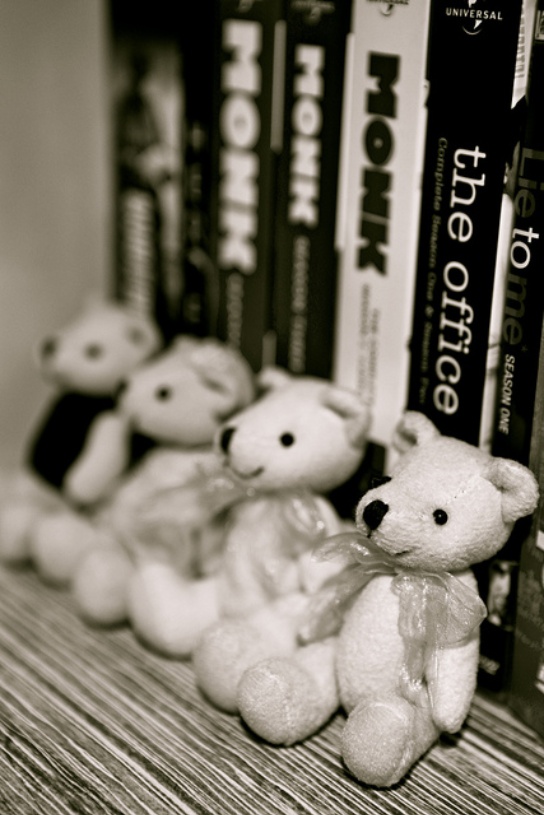}  &
    \includegraphics[width=5.85cm, keepaspectratio]{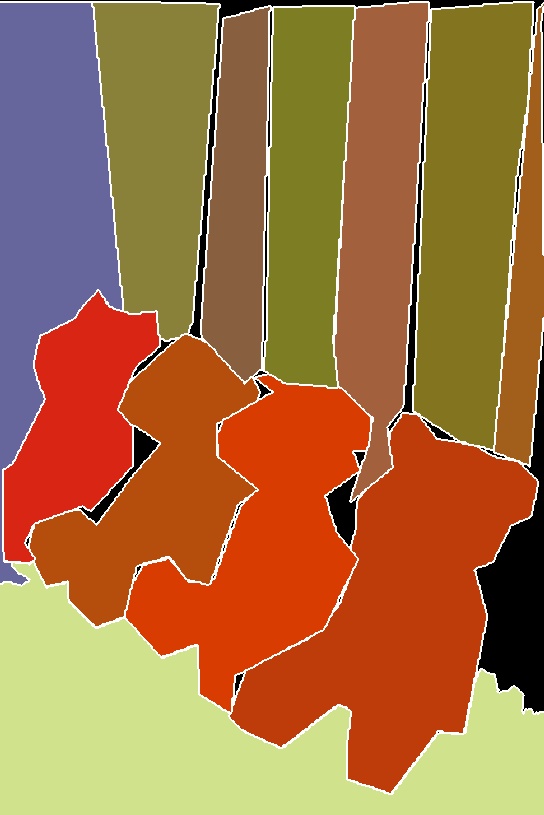}  &
    \includegraphics[width=5.85cm, keepaspectratio]{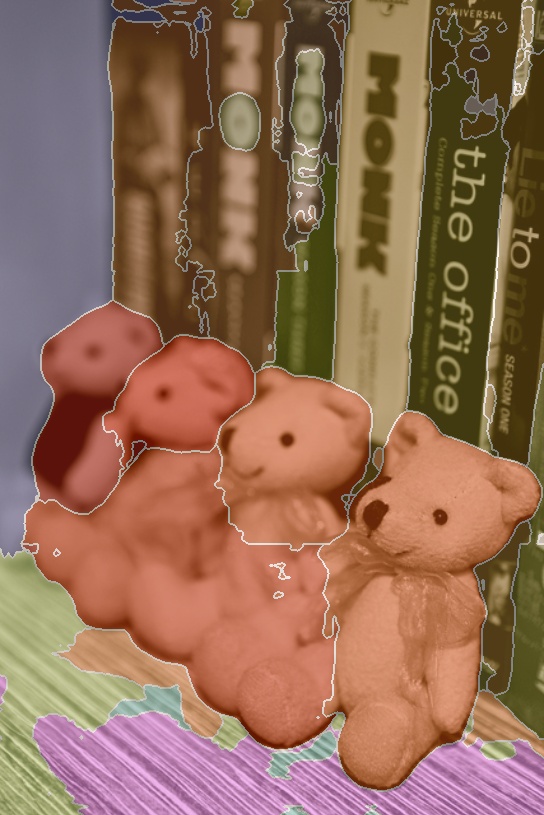} \\
 \end{tabular}
  \caption{Failure cases on COCO dataset.}
  \label{fig:coco_failure}
\end{figure*}
\endgroup

{\small
\bibliographystyle{ieee_fullname}
\bibliography{egbib}

\begin{thebibliography}{10}\itemsep=-1pt

\bibitem{chen2018masklab}
Liang-Chieh Chen, Alexander Hermans, George Papandreou, Florian Schroff, Peng
  Wang, and Hartwig Adam.
\newblock Masklab: Instance segmentation by refining object detection with
  semantic and direction features.
\newblock In {\em Proceedings of the IEEE Conference on Computer Vision and
  Pattern Recognition}, pages 4013--4022, 2018.

\bibitem{chen2017rethinking}
Liang-Chieh Chen, George Papandreou, Florian Schroff, and Hartwig Adam.
\newblock Rethinking atrous convolution for semantic image segmentation.
\newblock {\em arXiv preprint arXiv:1706.05587}, 2017.

\bibitem{chen2015mxnet}
Tianqi Chen, Mu Li, Yutian Li, Min Lin, Naiyan Wang, Minjie Wang, Tianjun Xiao,
  Bing Xu, Chiyuan Zhang, and Zheng Zhang.
\newblock Mxnet: A flexible and efficient machine learning library for
  heterogeneous distributed systems.
\newblock {\em arXiv preprint arXiv:1512.01274}, 2015.

\bibitem{cordts2016cityscapes}
Marius Cordts, Mohamed Omran, Sebastian Ramos, Timo Rehfeld, Markus Enzweiler,
  Rodrigo Benenson, Uwe Franke, Stefan Roth, and Bernt Schiele.
\newblock The cityscapes dataset for semantic urban scene understanding.
\newblock In {\em Proceedings of the IEEE conference on computer vision and
  pattern recognition}, pages 3213--3223, 2016.

\bibitem{de2017semantic}
Bert De~Brabandere, Davy Neven, and Luc Van~Gool.
\newblock Semantic instance segmentation with a discriminative loss function.
\newblock {\em arXiv preprint arXiv:1708.02551}, 2017.

\bibitem{de2018panoptic}
Daan de Geus, Panagiotis Meletis, and Gijs Dubbelman.
\newblock Panoptic segmentation with a joint semantic and instance segmentation
  network.
\newblock {\em arXiv preprint arXiv:1809.02110}, 2018.

\bibitem{dumoulin2017learned}
Vincent Dumoulin, Jonathon Shlens, and Manjunath Kudlur.
\newblock A learned representation for artistic style.
\newblock {\em Proc. of ICLR}, 2, 2017.

\bibitem{fathi2017semantic}
Alireza Fathi, Zbigniew Wojna, Vivek Rathod, Peng Wang, Hyun~Oh Song, Sergio
  Guadarrama, and Kevin~P Murphy.
\newblock Semantic instance segmentation via deep metric learning.
\newblock {\em arXiv preprint arXiv:1703.10277}, 2017.

\bibitem{ganin2016deepwarp}
Yaroslav Ganin, Daniil Kononenko, Diana Sungatullina, and Victor Lempitsky.
\newblock Deepwarp: Photorealistic image resynthesis for gaze manipulation.
\newblock In {\em European Conference on Computer Vision}, pages 311--326.
  Springer, 2016.

\bibitem{ghiasi2017exploring}
Golnaz Ghiasi, Honglak Lee, Manjunath Kudlur, Vincent Dumoulin, and Jonathon
  Shlens.
\newblock Exploring the structure of a real-time, arbitrary neural artistic
  stylization network.
\newblock {\em arXiv preprint arXiv:1705.06830}, 2017.

\bibitem{he2017mask}
Kaiming He, Georgia Gkioxari, Piotr Doll{\'a}r, and Ross Girshick.
\newblock Mask r-cnn.
\newblock In {\em Proceedings of the IEEE international conference on computer
  vision}, pages 2961--2969, 2017.

\bibitem{huang2017arbitrary}
Xun Huang and Serge Belongie.
\newblock Arbitrary style transfer in real-time with adaptive instance
  normalization.
\newblock In {\em Proceedings of the IEEE International Conference on Computer
  Vision}, pages 1501--1510, 2017.

\bibitem{karras2018style}
Tero Karras, Samuli Laine, and Timo Aila.
\newblock A style-based generator architecture for generative adversarial
  networks.
\newblock {\em arXiv preprint arXiv:1812.04948}, 2018.

\bibitem{kirillov2019panoptic}
Alexander Kirillov, Ross Girshick, Kaiming He, and Piotr Doll{\'a}r.
\newblock Panoptic feature pyramid networks.
\newblock {\em arXiv preprint arXiv:1901.02446}, 2019.

\bibitem{kirillov2018panoptic}
Alexander Kirillov, Kaiming He, Ross Girshick, Carsten Rother, and Piotr
  Doll{\'a}r.
\newblock Panoptic segmentation.
\newblock {\em arXiv preprint arXiv:1801.00868}, 2018.

\bibitem{kong2018recurrent}
Shu Kong and Charless~C Fowlkes.
\newblock Recurrent pixel embedding for instance grouping.
\newblock In {\em Proceedings of the IEEE Conference on Computer Vision and
  Pattern Recognition}, pages 9018--9028, 2018.

\bibitem{li2018learning}
Jie Li, Allan Raventos, Arjun Bhargava, Takaaki Tagawa, and Adrien Gaidon.
\newblock Learning to fuse things and stuff.
\newblock {\em arXiv preprint arXiv:1812.01192}, 2018.

\bibitem{li2018weakly}
Qizhu Li, Anurag Arnab, and Philip~HS Torr.
\newblock Weakly-and semi-supervised panoptic segmentation.
\newblock In {\em Proceedings of the European Conference on Computer Vision
  (ECCV)}, pages 102--118, 2018.

\bibitem{li2018attention}
Yanwei Li, Xinze Chen, Zheng Zhu, Lingxi Xie, Guan Huang, Dalong Du, and
  Xingang Wang.
\newblock Attention-guided unified network for panoptic segmentation.
\newblock {\em arXiv preprint arXiv:1812.03904}, 2018.

\bibitem{lin2017focal}
Tsung-Yi Lin, Priya Goyal, Ross Girshick, Kaiming He, and Piotr Doll{\'a}r.
\newblock Focal loss for dense object detection.
\newblock In {\em Proceedings of the IEEE international conference on computer
  vision}, pages 2980--2988, 2017.

\bibitem{lin2014microsoft}
Tsung-Yi Lin, Michael Maire, Serge Belongie, James Hays, Pietro Perona, Deva
  Ramanan, Piotr Doll{\'a}r, and C~Lawrence Zitnick.
\newblock Microsoft coco: Common objects in context.
\newblock In {\em European conference on computer vision}, pages 740--755.
  Springer, 2014.

\bibitem{liu2018intriguing}
Rosanne Liu, Joel Lehman, Piero Molino, Felipe~Petroski Such, Eric Frank, Alex
  Sergeev, and Jason Yosinski.
\newblock An intriguing failing of convolutional neural networks and the
  coordconv solution.
\newblock In {\em Advances in Neural Information Processing Systems}, pages
  9628--9639, 2018.

\bibitem{liu2018path}
Shu Liu, Lu Qi, Haifang Qin, Jianping Shi, and Jiaya Jia.
\newblock Path aggregation network for instance segmentation.
\newblock In {\em Proceedings of the IEEE Conference on Computer Vision and
  Pattern Recognition}, pages 8759--8768, 2018.

\bibitem{neuhold2017mapillary}
Gerhard Neuhold, Tobias Ollmann, Samuel Rota~Bulo, and Peter Kontschieder.
\newblock The mapillary vistas dataset for semantic understanding of street
  scenes.
\newblock In {\em Proceedings of the IEEE International Conference on Computer
  Vision}, pages 4990--4999, 2017.

\bibitem{novotny2018semi}
David Novotny, Samuel Albanie, Diane Larlus, and Andrea Vedaldi.
\newblock Semi-convolutional operators for instance segmentation.
\newblock In {\em Proceedings of the European Conference on Computer Vision
  (ECCV)}, pages 86--102, 2018.

\bibitem{xie2017aggregated}
Saining Xie, Ross Girshick, Piotr Doll{\'a}r, Zhuowen Tu, and Kaiming He.
\newblock Aggregated residual transformations for deep neural networks.
\newblock In {\em Proceedings of the IEEE Conference on Computer Vision and
  Pattern Recognition}, pages 1492--1500, 2017.

\bibitem{xiong2019upsnet}
Yuwen Xiong, Renjie Liao, Hengshuang Zhao, Rui Hu, Min Bai, Ersin Yumer, and
  Raquel Urtasun.
\newblock Upsnet: A unified panoptic segmentation network.
\newblock {\em arXiv preprint arXiv:1901.03784}, 2019.

\bibitem{yang2019deeperlab}
Tien-Ju Yang, Maxwell~D Collins, Yukun Zhu, Jyh-Jing Hwang, Ting Liu, Xiao
  Zhang, Vivienne Sze, George Papandreou, and Liang-Chieh Chen.
\newblock Deeperlab: Single-shot image parser.
\newblock {\em arXiv preprint arXiv:1902.05093}, 2019.

\bibitem{zhang2019bag}
Zhi Zhang, Tong He, Hang Zhang, Zhongyuan Zhang, Junyuan Xie, and Mu Li.
\newblock Bag of freebies for training object detection neural networks.
\newblock {\em arXiv preprint arXiv:1902.04103}, 2019.

\bibitem{zhao2017pyramid}
Hengshuang Zhao, Jianping Shi, Xiaojuan Qi, Xiaogang Wang, and Jiaya Jia.
\newblock Pyramid scene parsing network.
\newblock In {\em Proceedings of the IEEE conference on computer vision and
  pattern recognition}, pages 2881--2890, 2017.

\end{thebibliography}
}

\end{document}